\documentclass[runningheads]{llncs}

\usepackage{eccv}

\usepackage{eccvabbrv}
\usepackage{enumerate}
\usepackage{graphicx}
\usepackage{booktabs}
\usepackage{makecell}
\usepackage{multirow}
\usepackage{wrapfig}

\usepackage{tikz}
\usetikzlibrary{shapes,arrows,positioning}

\usepackage[accsupp]{axessibility}  

\usepackage{hyperref}

\usepackage{orcidlink}

\begin{document}

\title{Retargeting Visual Data with Deformation Fields}


\author{Tim Elsner\inst{1}\orcidlink{0000-0003-1309-8003} \and
Julia Berger\inst{1}\orcidlink{0009-0002-8086-2788} \and
Tong Wu\inst{2}\orcidlink{0000-0001-9974-3821} \and
Victor Czech\inst{1}\orcidlink{0009-0008-0918-7998} \and 
Lin Gao\inst{2}\and  
Leif Kobbelt\inst{1}\orcidlink{0000-0002-7880-9470}
}
%
\authorrunning{T. Elsner, J. Berger et al.}

\institute{RWTH Aachen University, Germany
\email{\{elsner,czech,kobbelt\}@cs.rwth-aachen.de / julia.berger@rwth-aachen.de} \and
University of Chinese Academy of Sciences, China\\
\email{\{wutong19s,gaolin\}@ict.ac.cn}}

\maketitle

\begin{abstract}
Seam carving is an image editing method that enables content-aware resizing, including operations like removing objects. 
However, the seam-finding strategy based on dynamic programming or graph-cut limits its applications to broader visual data formats and degrees of freedom for editing.
Our observation is that describing the editing and retargeting of images more generally by a \emph{deformation field} yields a generalisation of content-aware deformations. 
We propose to learn a deformation with a neural network that keeps the output plausible while trying to deform it only in places with low information content. This technique applies to different kinds of visual data, including images, 3D scenes given as neural radiance fields, or even polygon meshes. Experiments conducted on different visual data show that our method achieves better content-aware retargeting compared to previous methods.
\end{abstract}
\section{Introduction}
\begin{figure}[h]
    \centering
    \begin{minipage}{0.55\textwidth}
        \centering
        \includegraphics[width=\textwidth]{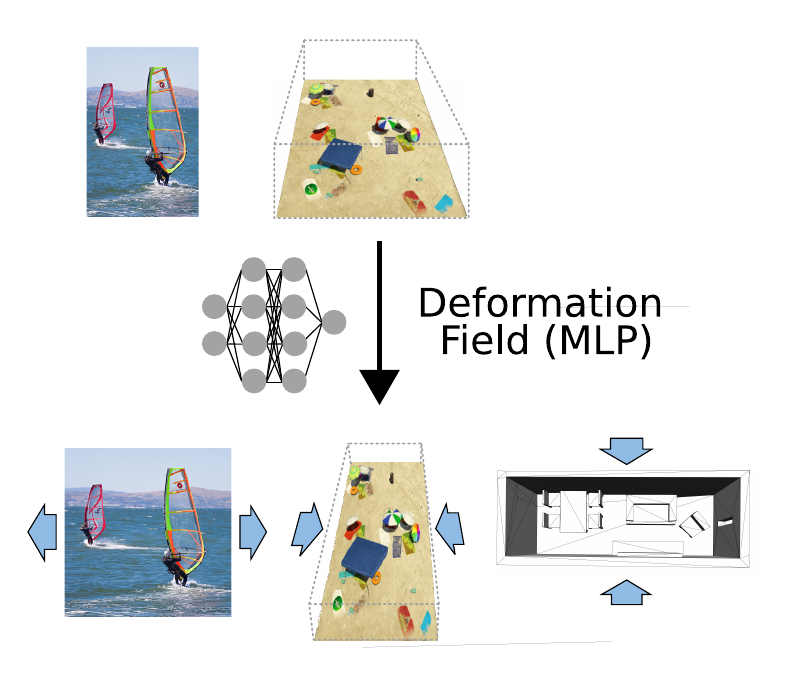} 
        \caption{Different objectives applied to different visual domains with our approach: We demonstrate retargeting images, NeRFs, and meshes. 'Surfer' image from~\cite{retargetme}.}
        \label{fig:teaser}
    \end{minipage}\hfill
    \begin{minipage}{0.35\textwidth}
        \begin{tikzpicture}[
            scale=0.1,
            node distance=0.2cm and 0.2cm,
            ar/.style={->,>=latex},
            mynode/.style={
                draw,
                text width=1.2cm,
                align=center
            }
        ]
            \node[mynode][font=\tiny] (imres) {Image Resizing};
            \node[mynode, left=of imres][font=\tiny] (crop) {Cropping};
            \node[mynode, below left=of imres][font=\tiny] (roi) {ROI-based, \eg \cite{LiuG05}};
            \node[mynode, below right=of roi][font=\tiny] (retar) {Retargeting};
            \node[mynode, right=of imres][font=\tiny] (scale) {Naive Scaling};
            
            \node[mynode, below left=of retar][font=\tiny] (disc) {Discrete};
            \node[mynode, below right=of retar][font=\tiny] (cont) {Continuous};
            \node[mynode, below right=of disc][font=\tiny] (seam) {Seam Carving\cite{classic_seam} };
            \node[mynode, below=of disc][font=\tiny] (others) {Others \cite{rl_seamcarve, tan2019cycle}}; 
            \node[mynode, below=of cont, dashed, red][font=\tiny] (ours) {Ours,\\Deformation};
            
            \node[mynode, below=of seam.south, text width=3cm][font=\tiny] (ext) {\textbf{Extensions:}\\ \eg.Video \cite{video_seam},\\better energy terms, etc.~ \cite{Dong09, Dong12, DongZLWKZ14, RubinsteinSA09, WuWFWLH10, SongLL19} }; 
            
            \draw[ar] (imres.south) -| (retar);
            \draw[ar] (imres) -- (roi);
            \draw[ar] (imres) -- (crop);
            \draw[ar] (imres) -- (scale);
            
            \draw[ar] (retar) -- (disc);
            \draw[ar] (retar) -- (cont);
            \draw[ar] (disc) -- (others);
            \draw[ar] (disc) -- (seam);
            \draw[ar, dashed, red] (cont.south) -| (ours);
            \draw[ar,] (seam.south) -- (ext);
            
            \draw[ar, dashed, red] (ours) -- (ext);
            
        \end{tikzpicture}
        \caption{The family of image resizing methods. Our approach formulates the problem in a more general way, but can utilise \eg new energy formulations as well.}
        \label{fig:taxonomy}
    \end{minipage}
\end{figure}

\label{sec:intro}
Media retargeting is a way to edit images, videos, 3D objects, or even entire 3D scenes by a global deformation such that relevant content, details, and features are properly preserved. The overall goal is to change the aspect ratio of the bounding box of the data so that it fits an allocated space, \eg fitting an image to a display with prescribed format. Retargeting is based on the idea of identifying regions with little detail and to accumulate the necessary distortion induced by the deformation in these regions. Small editing operations are possible \eg through concentrating the distortion there, specifying a region containing a particular object to have this object removed.\\
Most existing retargeting methods typically use some form of seam carving, \ie the deletion of discrete pixels on a path through the image (\textit{seams})~\cite{classic_seam,RubinsteinSA09, Dong09, Dong12}. 
As previous methods perform a discrete removal or addition of pixels, they do not natively extend to \eg continuous domains like Neural Radiance Fields~\cite{mildenhall2020nerf} (NeRFs) and are often restricted to a class of solutions defined by the used algorithm.\\
Our approach formulates the problem via continuous deformation fields, mapping from the retargeted data back to the undistorted input. It natively extends from images on other domains, like NeRFs and polygon meshes (\cref{fig:teaser}), and offers better solutions through a \emph{global} and \emph{continuous} instead of a discrete and greedy optimisation process (see \cref{fig:example_non_45deg}). Our contributions are as follows:
\begin{itemize}
\item We introduce the use of a \emph{neural deformation field} to compress or stretch unimportant regions to achieve a smaller or larger output or to follow other editing objectives.
\item We abstract and generalise the underlying rules of creating plausible outputs with seam carving, then demonstrate the domain agnostic nature of our formulation by applying it to images, 3D meshes, and 3D scenes in the form of NeRFs.
\item We show that the high flexibility of our approach from optimising a global deformation field produces better outputs than iteratively computing seams, while also allowing configurations that are not possible with previous methods.
\end{itemize}
To achieve this, we regularise the neural deformation fields to follow \emph{general sanity guidelines}, while attempting to minimise distortion in places with presumed high information content. We evaluate our results on both qualitative and quantitative levels. To showcase our approach on interesting scenes captured as NeRFs and inspired by the dataset of Richter \etal~\cite{gtav_intel}, we also provide a small synthetic NeRF retargeting dataset captured in the video game \textit{GTA V}~\cite{GTA} by flying the camera over the virtual city, capturing the screen content and camera parameters.\\
Our method is a more reliable backbone for approaches following seam carving while outperforming current methods, requires only minor modifications for vastly different applications and domains, and is straightforward to implement. We provide the code of our method demonstrated on images in the supplemental.
\phantom{\cite{SongLL19}}
\section{Related Work}
As our approach is re-formulating \textit{seam carving}, it produces results from a \textit{single example} without any learned prior, and is designed to allow seam carving-like \textit{deformations for (learned) 3D representations}. We hence discuss these three branches of related work. 
\paragraph{Single Example Generation}
The goal of single example generation is to discover similar local patterns from a single input and synthesise novel samples. 
Texture synthesis on 2D images and 3D shapes has been widely studied in computer vision and computer graphics by non-neural methods~\cite{EfrosL99, WeiL00, EfrosF01, KFCODLW07}. 
There are also methods focusing on synthesising geometric textures for 3D shapes~\cite{lai2005geometric, Berkiten17, hertz2020deep}, which transfer local geometry details from an example shape to other surfaces. 
Recently, neural-based methods have been proposed to generate 2D textures~\cite{GatysEB15, Xian_2018_CVPR, ShahamDM19} and transfer them to 3D shapes~\cite{henzler2020neuraltexture,HenzlerDMR21}, resulting in improved results. 
With recent advances in implicit representation~\cite{IMNET, DeepSDF, OccNet} and neural radiance fields~\cite{mildenhall2020nerf}, Wu \etal propose to learn implicit 3D shapes from a shape with~\cite{{wu2022learningtexture}} or without texture~\cite{wu2022learning}. 
To enhance the realism of synthesised shapes, Huang \etal~\cite{NeRF-Texture} reconstruct the 3D shape with a NeRF by optimising texture features on a reconstructed mesh. These features can be applied to arbitrary surfaces to render the target surface with synthesised textures. These methods often only work for a certain kind of data format, while our method proposes a domain agnostic approach to synthesise visual data from a single example. 
\paragraph{Deformations for Learned 3D Representations}
Neural radiance fields (NeRF), as a newly proposed 3D shape representation, has been widely used in scene reconstruction and novel view synthesis. 
However, the original NeRF only works for static scenes, thus is unable to deal with deformation in dynamic scenes or generate motions for a static scene. 
To overcome this, the deformation field is introduced to NeRF. 
NeRF-Editing~\cite{NeRFEditing} is the first to explore deformation in NeRF. 
It first reconstructs the explicit mesh of a static scene with NeuS~\cite{NeuS} and deforms the explicit mesh with the ARAP algorithm~\cite{ARAP}. 
Then sample points in volume rendering are deformed along with the mesh via barycentric coordinate interpolation, resulting in the change of rendered images. 
Follow-up NeRF deformation methods~\cite{xu2022deforming, peng2021CageNeRF, VolTeMorph} employ a similar pipeline but use different geometry proxies. 
To reconstruct dynamic scenes, Albert \etal~\cite{D-NeRF} first explore the possibility by adding an extra time-conditioned deformation field to the static NeRF. 
The time-conditioned deformation field transforms sample points in the observation space into the canonical space, where their colours and opacities are queried and then rendered into images. 
This idea stimulated a series of dynamic NeRF reconstruction methods~\cite{Nerfies, HyperNeRF, Cai2022NDR, luiten2023dynamic, yang2023deformable3dgs, wu20234dgaussians}. 
While our approach also uses a deformation field, we inject general plausibility constraints into it to keep the results reasonable, enabling the idea behind seam carving on different types of visual data.

\paragraph{Seam Carving}
Image/video retargeting is an important tool in computer vision and computer graphics. 
Early works~\cite{LiuG05} first annotate or detect the region of interest~(ROI), which is uniformly scaled to the target size afterwards. 
While contents out of the ROI are deformed by fisheye warping, Liu and Gleicher~\cite{LiuG06} later extend it to the video domain by introducing a motion salience map. 
Avidan and Shamir~\cite{classic_seam} propose seam carving, a content-aware image resizing method. 
It first defines a saliency map by summing the gradient norms along the horizontal and vertical directions. The cost of a seam is defined by adding up the values on the salience map (energy). 
The seam with minimal energy can be found using dynamic programming and will be removed in the resizing process. 
Rubinstein \etal~\cite{video_seam} extend the seam carving idea to the video domain and transform the dynamic programming problem to a graph cut problem. 
Follow-up works~\cite{RubinsteinSA09, Dong09, Dong12} incorporate homogeneous cropping and scaling with seam carving and excavate more high-level information like symmetry~\cite{WuWFWLH10}, depth~\cite{BashaMA13}, and object similarity~\cite{DongZLWKZ14} to assist seam carving. 
With the advances in deep learning, a few methods~\cite{MoreiraSPPC22, NamAYKSL21, abs-2109-01764} made attempts to detect and localise the seam carving forgery. 
Song \etal~\cite{SongLL19} propose Carving-Net to replace the previous handcrafted energy map with a neural network while the seam finding algorithm is still dynamic programming. \\~\\
Generally, our method is invariant to the underlying energy term, and hence can be replaced by any other formulation like the ones mentioned before. This relationship is visualised in the taxonomy in \cref{fig:taxonomy}: Our method solves a global, continuous optimisation problem instead of greedy and local steps formulated as dynamic programming, and can hence function as a more flexible backbone for approaches that derive from seam carving. This formulation allows more complex solutions that work better in many scenarios (see \cref{fig:example_non_45deg}), while the space of all possible outputs contains the solutions of seam carving.
\begin{figure*}[h]
\centering
\includegraphics[width=1.0\textwidth]{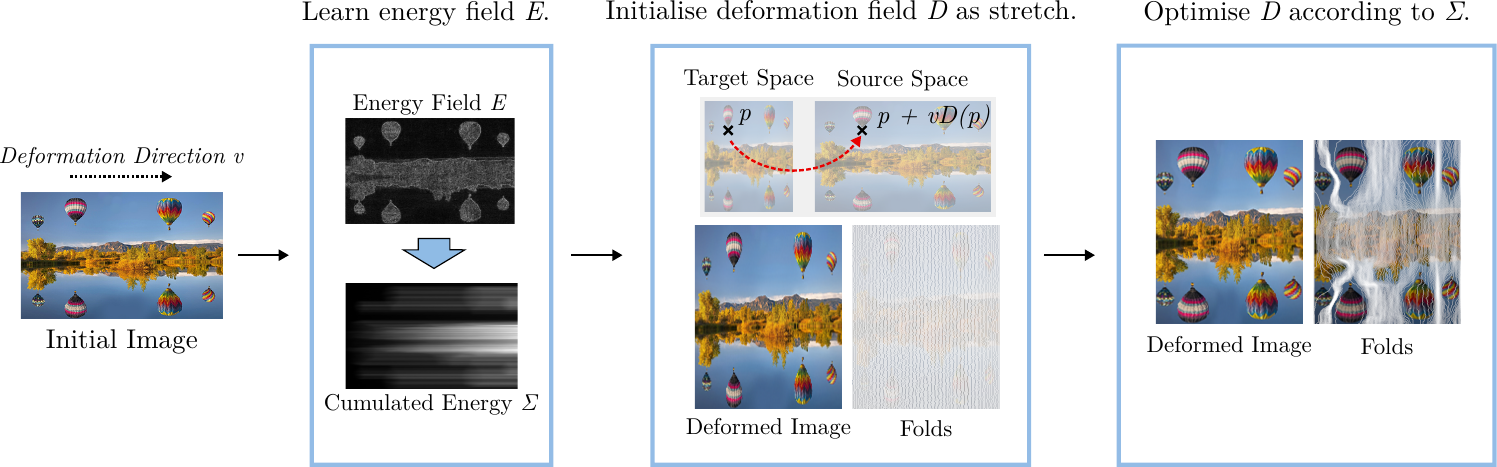}
\caption{Our proposed pipeline for retargeting visual data: For a given input (left), we train two simple networks that learn the energy and cumulative energy along the deformation axis of the input (centre-left). We initialise a network that stretches samples to the desired position (centre-right), then optimise this deformation to distribute the distortion to low information content regions (right). Balloon image from \cite{retargetme}.}\label{fig:pipeline_general}\label{fig:pipeline_images}
\end{figure*}
\begin{figure*}[h]
\centering
\includegraphics[width=1.0\textwidth]{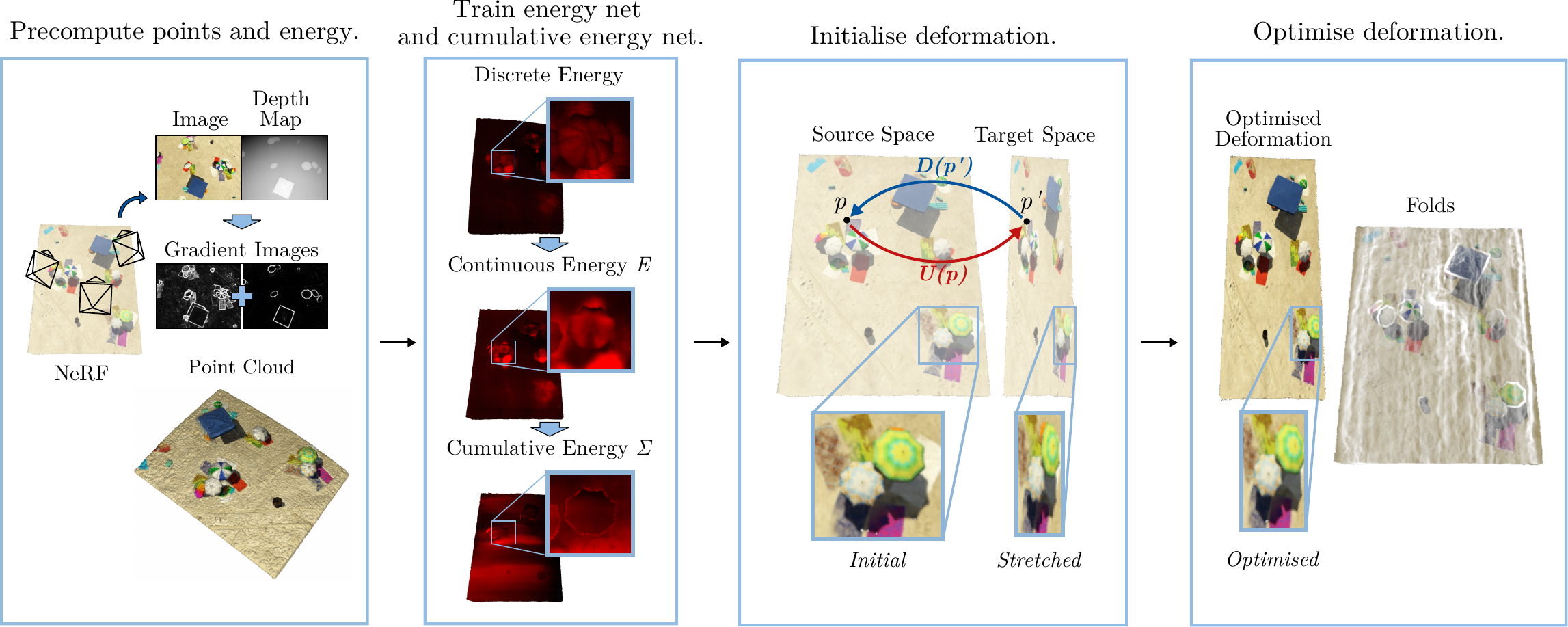}
\caption{Our pipeline applied to neural radiance fields: We first obtain a point cloud, then estimate the energy values. We then learn a continuous energy function and a cumulative energy function that we use to optimise our deformation network. The resulting deformation field is visualised on the bottom right.}\label{fig:pipeline_nerf}
\end{figure*}

\begin{figure}[htpb]
	\centering
	{
		\begin{tabular}{ccccc}
                \includegraphics[height=0.32\linewidth]{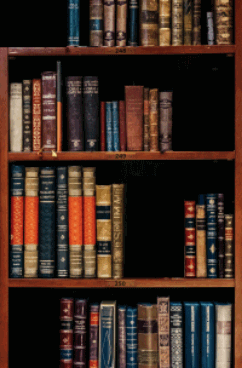}&
                \includegraphics[height=0.32\linewidth]{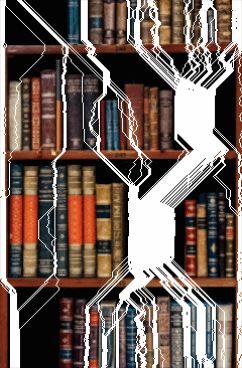}&
                \includegraphics[height=0.32\linewidth]{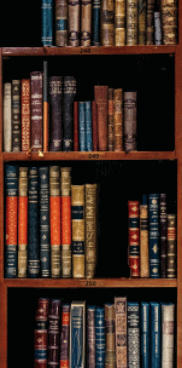}&
                \includegraphics[height=0.32\linewidth]{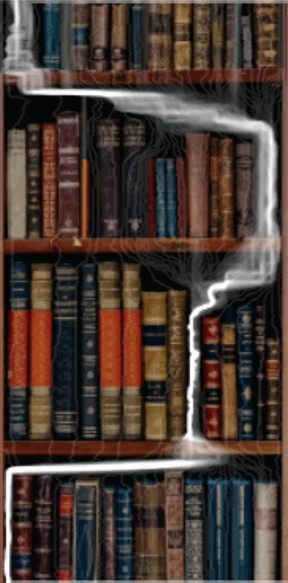}&
                \includegraphics[height=0.32\linewidth]{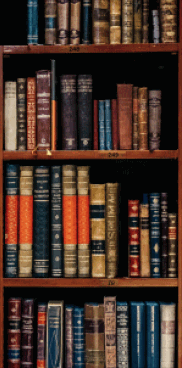}
                \\
                Input &
                Seams to carve &
                Seam carved &
                Our folds &
                Ours
		\end{tabular}
	}
	\caption{Example of applying retargeting to $75\%$ width. Seam carving fails to use seams that are non-continuous or with a too steep angle, not properly using the empty space in the shelf to close the gaps before shrinking the bookshelves content.}
    \label{fig:example_non_45deg}
\end{figure}
\begin{figure}[h]
\centering
    \centering
    \begin{tabular}{ccc}
        \includegraphics[height=0.25\linewidth]{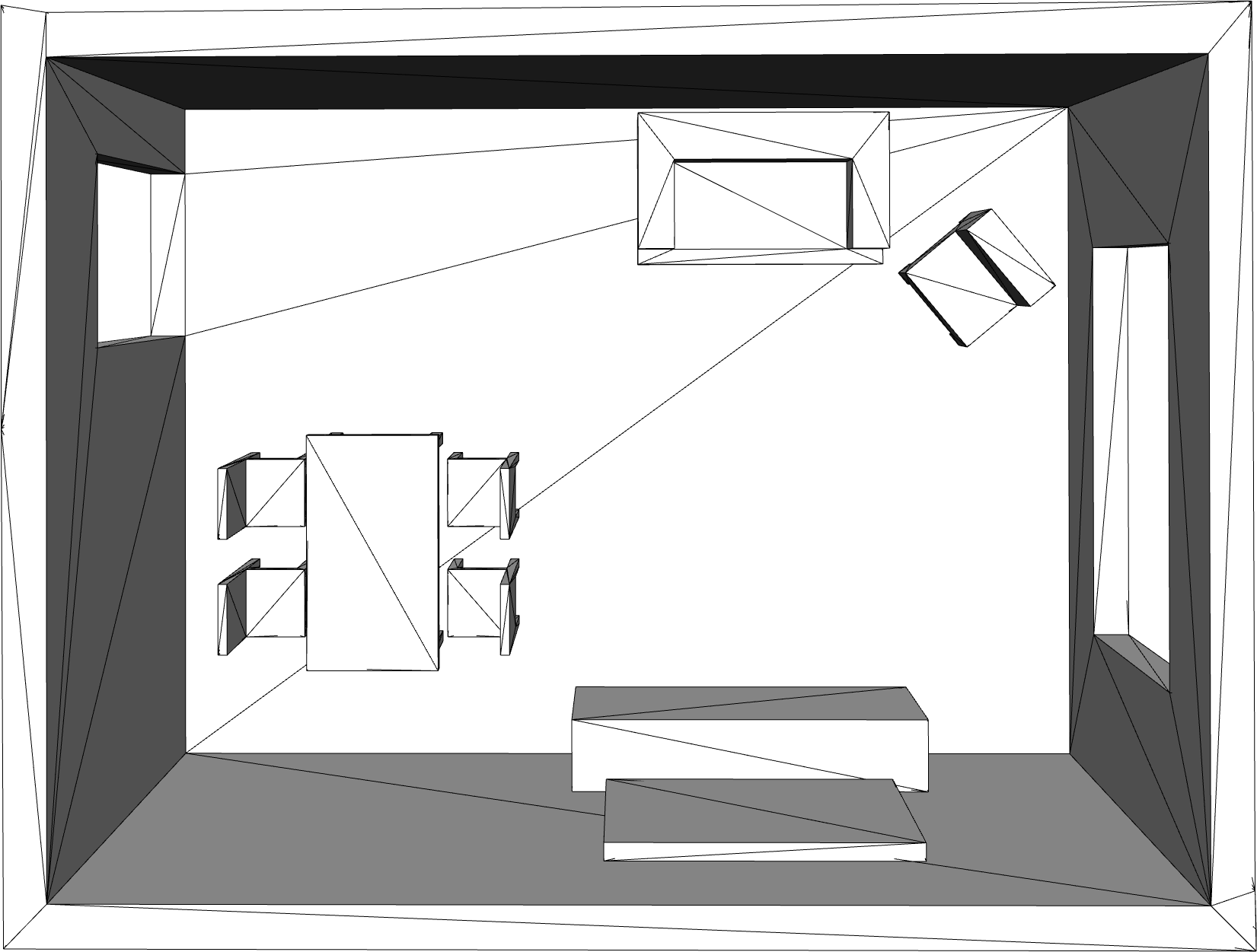}&
        \includegraphics[height=0.25\linewidth]{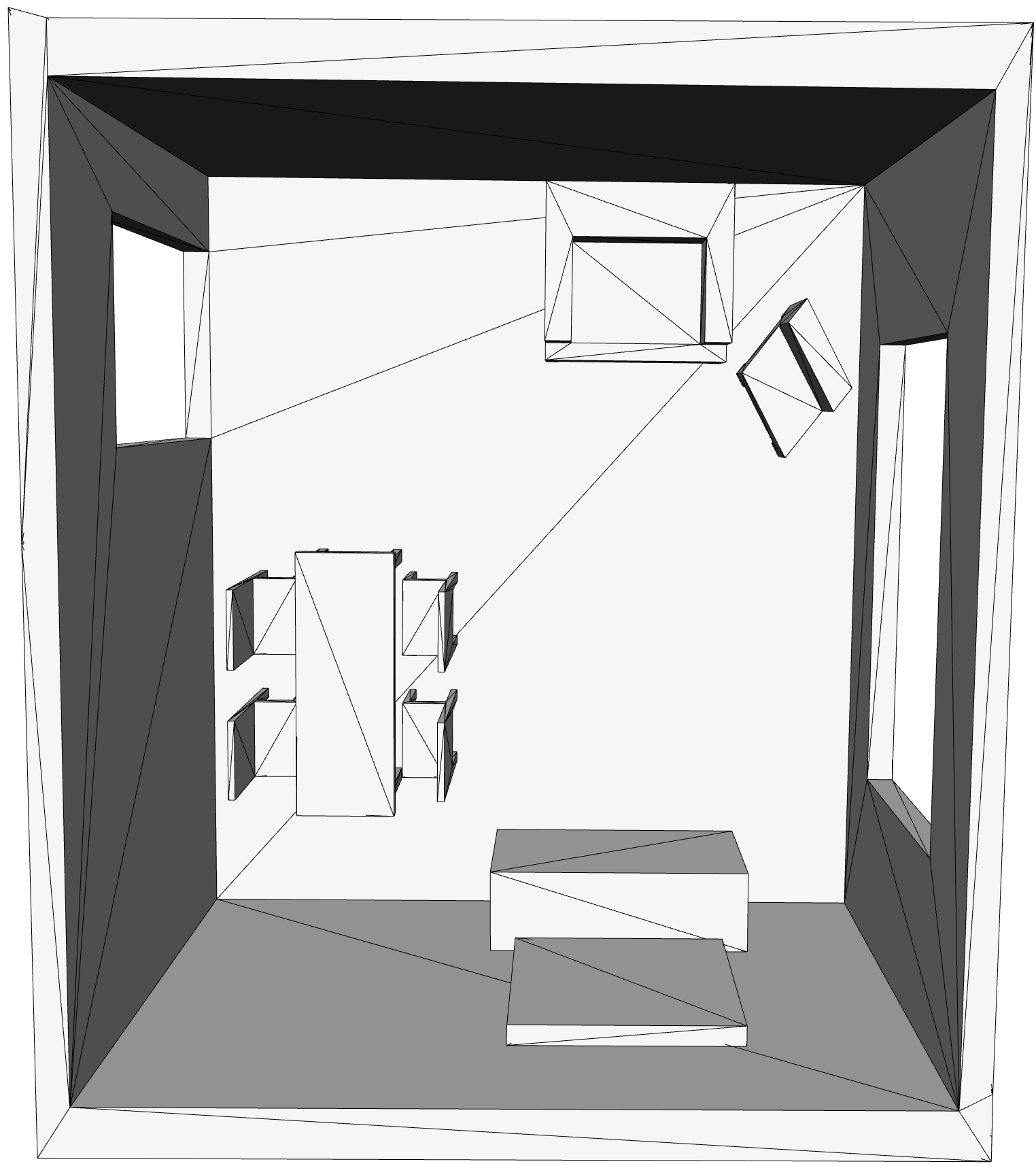}&
        \includegraphics[height=0.25\linewidth]{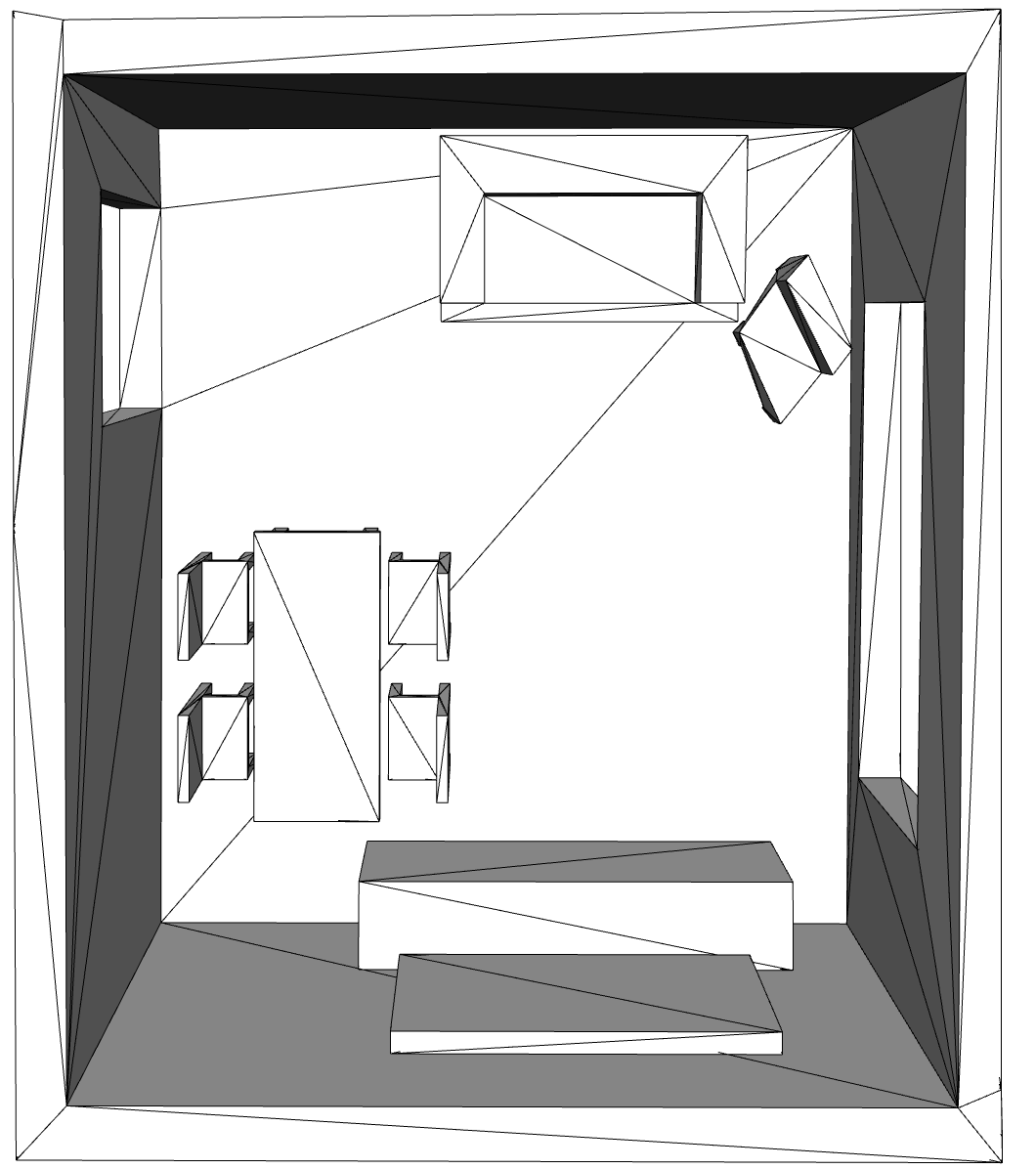}
        \\
        Input &
        Uniform &
        Optimised
    \end{tabular}
\caption{Applying our retargeting to 3D meshes, moving furniture closer together to retarget the scene to a smaller size while limiting distortion to objects.}\label{fig:seam_mesh}
\end{figure}


\section{Deformation Fields for Retargeting Visual Data}\label{sec:main_sec}
Editing strategies for visual data require avoiding changes to meaningful content, that create visible artefacts.
An approach that focuses on retargeting the input to match a target size can be guided by the following set of rules:
\begin{itemize}
\item \textit{Content Aware Objective}: Prevent the accidental creation of new or the removal of existing regions with high information content (control over \textbf{where} deformation happens).
\item \textit{Sanity Objective}: Avoid introducing deformations that yield implausible results (control over \textbf{how} deformation happens).
\end{itemize}\label{enum:criteria}

In the following sections, we are only concerned with the retargeting to a smaller size (\textit{shrinking}) for ease of explanation. The concept can naturally be reformulated to expand the visual data instead (\textit{expansion}, see Sec.~ C (Appendix)) or to include editing (see \cref{sec:editing}). 
\\
Seam carving~\cite{classic_seam} and its derivatives produce a \textit{discrete solution} to this by performing edits through manipulating (removing or repeating) one seam of pixels at a time, which is a single-pixel-wide path passing through the image. Through dynamic programming, seam carving chooses this path based on the lowest energy of the pixels, with high energy indicating information-rich areas. In a simple case, this energy is the colour gradient. 
This yields a discrete solution to the retargeting problem that is both content aware and plausible.  

To obtain a \textit{continuous solution}, instead of extracting discrete seams of the same width, we use continuous \emph{folds} of varying width using a learned continuous deformation function. This formulation optimises a solution that is a superset of all existing seam carving solutions (see \cref{fig:example_non_45deg}).
We always use the same core formulation, regardless of the underlying domain and application, \eg moving an object in an image or retargeting a 3D scene.
\subsection{General Formulation}\label{sec:general_formulation}
Our approach always optimises a deformation function that maps from points in the retargeted output space (\eg an image at half the size) $I: P \to \mathbb{R}^n$ to points in the original data $I': P' \to \mathbb{R}^n$ (\eg to $\mathbb{R}^3$ for RGB colours) by adding an offset to their coordinate. The value of a point $p$ in the output data $I$ is hence defined as the value of $p$ deformed to a new point $p'$ looked up in $I'$. To represent the deformation function $D$, we use a \emph{neural deformation field}, a simple MLP where $D: P \to \mathbb{R}$. To keep our approach simpler, we only allow scalar deformations along a fixed deformation direction $v$, \eg along the $x$ axis (width) of an image:
\begin{equation}
    I(p) = I'(p + D(p) \cdot v)
\end{equation}\label{eq:deformation}
Hence, to get the content for a point $p$ in the retargeted data, we offset the coordinate and look up the content at the offset coordinate $p'$ in the original data.
To fulfil the objectives stated in the beginning of \cref{sec:main_sec}, we introduce two sets of losses, content aware and sanity objectives:\\~\\
\textbf{Content Aware Objectives} apply regularisation to guide the deformation process when adjusting to the data content. In particular, we penalise deformation in regions with high information content to discourage noticeable changes. We define the \emph{energy function} $E: P' \to \mathbb{R}$ to be a measure for content information in the original visual data, expresses as, \eg, a colour gradient:
\begin{equation}\label{eq:energy}
    E(p') = ||\nabla I(p')||_2
\end{equation}
The overall \emph{content deformation} loss is the product of deformation magnitude at $p$ and energy at the original $p'$ over all $p$ in the deformed data:
\begin{equation}\label{eq:loss_energy}
    L_{C} = \int\limits_{p \in P}\left[E(p + v \cdot D(p)) \cdot ||\nabla D(p)||_1\right] \ dp
\end{equation}
Thus, in \cref{eq:loss_energy}, we effectively penalise visible deformations defined as the product of information content at points, namely $E(p + v \cdot D(p))$, and the deformation itself via $||\nabla D(p)||_1$. Using the $L^1$-norm for the distortion gradient promotes piecewise constant, possibly sparse, deformations.
We can then split up the change in deformation into change in deformation direction $v$ and change in direction $v^{\perp}$ orthogonal to $v$:
\begin{equation}\label{eq:rewrite}
\begin{aligned}
    L_{C} = \int\limits_{p \in P}\biggl[&E(p + v \cdot D(p)) \cdot \left(\left|\frac{\partial}{\partial v} D(p)\right| + \left|\frac{\partial}{\partial v^{\perp}} D(p)\right|\right)\biggr] \ dp
\end{aligned}
\end{equation}

This equation is then further split into two components, $L_e$ and $L_s$. The loss $L_e$ penalises \textit{stretching} or \textit{compression} and $L_s$ punishes \textit{shearing}, which we define to be deformation change in a direction that is not the deformation direction. For a now discrete sample set $P$ from our domain, these terms are expressed as:

\begin{equation}\label{eq:content}
\begin{aligned}
    L_e = & \sum_{p \in P}\left[E\left(p + v \cdot D(p)\right)\cdot \left|\frac{\partial}{\partial v} D(p)\right|\right] \\
    L_s = & \sum_{p \in P}\left[E\left(p + v \cdot D(p)\right)\cdot \left|\frac{\partial}{\partial v^{\perp}} D(p)\right|\right]
\end{aligned}
\end{equation}

Essentially, $L_e$ directs deformations to areas of low information content, while $L_s$ discourages the resulting shearing to affect areas of high information content. 
Note that while $L_s$ would also penalise non-straight deformation seams, we tune our losses such that this only impacts the output much if an area is sheared. An example of this is shown in \cref{fig:pipeline_general}: The deformations around the balloons introduce shearing only at the seam, in an area of low information content.

In case of image deformation, neighbouring pixels with low energy in the input may translate to pixels widely apart in the output, potentially bypassing a substantial amount of important content. Jumps in the deformation field $D$ indicate this effect. To prevent this, we penalise the amount of energy $\hat{E}$ \textit{between} two points $p', q' \in P'$ in the original domain, rather than the energy at individual positions, as in \cref{eq:energy}:
\begin{equation}\label{eq:energy_between}
    \hat{E}(p', q') = \int\limits_{r' \in [p', q']} E(r') \ dr'
\end{equation}

Using $\hat{E}$, we then redefine $L_e$ from \cref{eq:content} to consider energy and difference in deformation magnitude between $p$ and a slightly offset $p_\varepsilon := p + \varepsilon$ in deformation direction:

\begin{equation}\label{eq:content_final}
\begin{aligned}
    L_e = \sum_{p \in P} &\hat{E}\left(p + v D(p), p_{\varepsilon} + v D(p_{\varepsilon})\right) \cdot\frac{\left|D(p) - D(p_\varepsilon)\right|}{\varepsilon}
\end{aligned}
\end{equation}
This formulation guides the deformation to avoid introducing deformation to high information content regions or fold over them, while $L_s$ discourages non-local deformations that would apply shearing in high information areas. For example, shearing in blue sky is acceptable, shearing on a brick wall is not.~\\~\\
\textbf{Sanity objectives} prevent undesired outputs by introducing two additional losses. Assume we shrink the input by a factor $\alpha \in [0,1]$:
\begin{enumerate}\label{eq:loss_bound}
\item The boundaries of the affected region should not change, \eg an image should not be simply cropped. 
Based on $\alpha$, the deformation field should map the start/end of the deformation axis in the output to the start/end of the input. For $P_0$,  $P_1$ containing all points at the start/end of the deformation axis of the output, we express this requirement as follows:
\begin{equation}\label{eq:mono}
L_{b} = \int\limits_{p \in P_0} |D(p)| \ dp + \int\limits_{p \in P_1} |D(p) - (1 - \alpha)| \ dp
\end{equation}
\item We additionally demand \emph{monotonicity} in the deformation field, expressed as
\begin{equation}\label{eq:loss_mono}
L_{m} = \int\limits_{p \in P} \max\left(0, -\frac{\partial}{\partial v} D(p)\right) \ dp,
\end{equation}
indicating that any change in the deformation direction should be positive. Therefore, deformation $D$ should be monotonic and only stagnate or increase exclusively along $v$. This prevents the repetition of the same image portion, noticeable by jumps with inconsistent direction in the deformation field.
\end{enumerate}~\\
We then obtain the total loss $L$ with hyperparameters $\lambda_*$ as
\begin{equation}\label{eq:main}
    L = \lambda_e L_{e} + \lambda_s L_{s} + \lambda_b L_{b} + \lambda_m  L_{m} .
\end{equation}
We initialise the deformation function with a uniform stretch to the target size.

In the following, we demonstrate the application of this general formulation to different domains, while adapting the energy formulation to the problem. All implementation and architecture details are given in Sec.~ A (Appendix) and the exact loss terms in Sec.~ B (Appendix). 


\subsection{Optimising Deformation Fields for Images}\label{sec:images}
Instead of using image data directly, we train a continuous image representation MLP $I': P' \to \mathbb{R}^3$ to be our input data, that maps pixel positions to colours. This ensures consistency with our continuous approaches.\\

We train one MLP to learn the energy field $E$, holding the colour gradient per point as in \cref{eq:energy}. Using the learned $E$, we train another MLP to replicate $\hat{E}$ as a \emph{cumulative energy field} $\Sigma$. The difference in cumulative energy $\Sigma$ between two points $p',q' \in P'$ on the same axis $v$ is the absolute difference in information content between them. As this coincides with the definition of $\hat{E}$ from \cref{eq:energy_between}, we approximate the measure of content relevance between two points as:
\begin{equation}\label{eq:equiv}
    \hat{E}(p',q') \approx |\Sigma(q') - \Sigma(p')|
\end{equation}
This cumulative gradient formulation therefore simplifies the interval over $E$ in \cref{eq:energy_between} to a simple difference computation between two MLP queries: How much content is lost when the deformation skips a region.
The energy field $E$ can be trained to reproduce any energy formulation other than a colour gradient. This exchangeable energy formulation ensures compatibility with other retargeting approaches as shown in \cref{fig:taxonomy}. For all objectives, we use discrete approximations of the loss terms introduced in \cref{sec:general_formulation}. The whole pipeline is visualised in \cref{fig:pipeline_images}, and additional results can be found in Sec.~ G (Appendix). 


\subsection{Optimising Deformation Fields for Neural Radiance Fields}\label{sec:nerf}

We adapt our approach from \cref{sec:general_formulation} for 3D scenes given as trained NeRFs as follows (see \cref{fig:pipeline_nerf}): 
\begin{itemize}
    \item A trained NeRF provides colour and depth images, from which we extract surface samples. Each point is assigned an energy value defined by change in colour and depth at this point.
    \item Based on the surface samples and their energy values, we train a continuous energy field $E$ and a continuous cumulative energy field $\Sigma$, to easily access the amount of content between points.
    \item We use an \emph{inverse deformation network} $U$ to find points in the target space that correspond to surface points in the source space, keeping computing costs low by restricting our optimisation to surface points.
    \item We apply our regularisation terms on a mix of surface and random points to optimise our loss function.
\end{itemize}

\paragraph{1. Extracting a Sparse Set of Representatives and Energy} 
We extract a surface point cloud from the RGB and depth images rendered by NeRF at the training camera positions. Only these points are relevant for scene deformation. We then compute the energy value for each point corresponding to a pixel as a sum of colour change in the RGB image and depth change in the depth image. 
To reduce the impact of highly noisy energy values, we apply smoothing by selecting the minimum energy for each point within a neighbourhood radius of 5 points.

\paragraph{2. Defining an Energy Function} 
To obtain a continuous energy field, we train an auxiliary energy network $E$. The network maps all 3D points to their energy values, and clamps the energy to zero for points further than twice the average neighbour distance. The resulting energy field has zero energy in empty space and inside objects, and non-zero energy values for points near the surface.
Similar to our approach with images, we learn a cumulative energy network $\Sigma$. The network is trained to accumulate the energy values $E(p)$ from uniformly sampled points $p$ on a ray in deformation direction. We compute the energy for a segment between two points $p',q'\in \mathbb{R}^3$ along $v$ using their difference in cumulative energy $\Sigma(q')$ and $\Sigma(p')$ (\cf \cref{eq:equiv}).

\paragraph{3. Inverting the Deformation}\label{sec:inverse} We define an inverse deformation function to determine those points that lie on the surface \emph{after} deformation. In contrast to images, we only perform optimisation on a sparse set of points, namely surface points. As we optimise a deformation from the \textit{target space} to the \textit{source space}, we need the position of surface points in both spaces. By training an MLP $U$ to reverse the deformation $D$, \ie minimising $(U(D(x)) = x)^2$, we can effortlessly extract surface points, and hence their energy values, in the target space. This relation is also visualised in \cref{fig:pipeline_nerf}. We always update $U$ once after training one iteration of $D$.\\

\begin{figure}[h]
    \centering
    \begin{minipage}{0.5\textwidth}
	\centering
	{
			\setlength\tabcolsep{0pt}
			\begin{tabular*}{\columnwidth}{@{\extracolsep{\fill}} ccc }
				\hspace{-1mm}{\rotatebox{90}{\quad \hspace{2mm} {Input}}} &
				\hspace{-2mm}\includegraphics[height=0.35\textwidth]{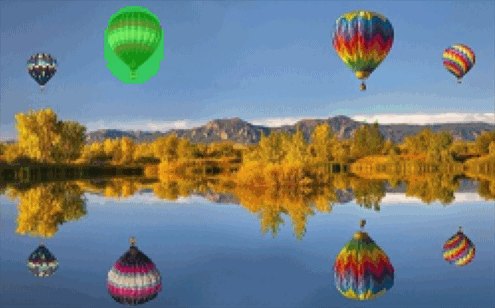}&
				\hspace{-2mm}\includegraphics[height=0.35\textwidth]{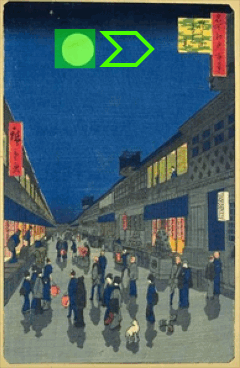}
				\\
				\hspace{-1mm}{\rotatebox{90}{\quad \hspace{2mm} {\makecell{Edited}}}} &
				\hspace{-2mm}\includegraphics[height=0.35\textwidth]{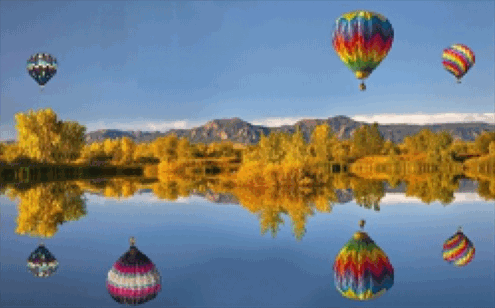}&
				\hspace{-2mm}\includegraphics[height=0.35\textwidth]{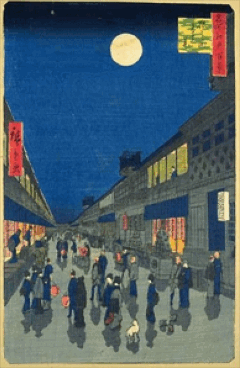}
			\end{tabular*}
	}
        \caption{We demonstrate different editing operations by adding and adapting the loss formulation. Left: Removing, right: Moving an object. Images from \cite{retargetme}.}
        \label{fig:editing}
    \end{minipage}\hfill
    \begin{minipage}{0.45\textwidth}
        \centering
            \centering
            \begin{subfigure}[b]{1.0\textwidth}
                \centering
                \includegraphics[width=\textwidth]{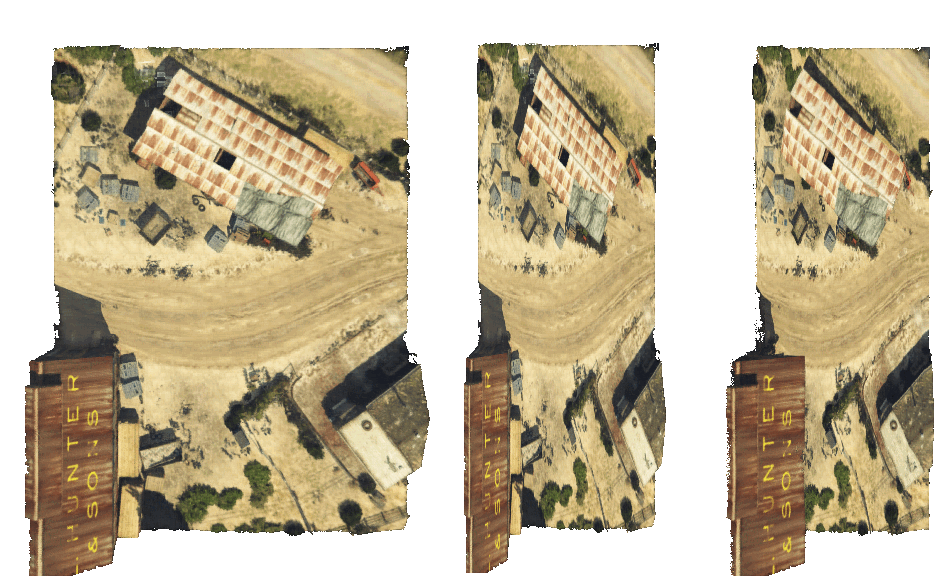}\\
                \includegraphics[width=\textwidth]{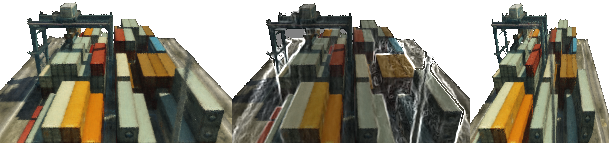}\\
            \end{subfigure}
        \caption{Applying our retargeting to NeRFs. Top row: Input, stretched, retargeted with our method. Bottom row: Input, folds (white) on the input suggested by our approach, retargeted with our method.}\label{fig:seam_nerfs}
    \end{minipage}
\end{figure}

As we use a discrete set of points, we apply the losses from \cref{sec:general_formulation} to \cref{eq:main} for a discrete interval. In particular, we average the loss over a random subset of sample points. Example images from deformed scenes can be found in \cref{fig:seam_nerfs}, while a larger number of results can be found in Sec.~ G (Appendix).

\subsection{Optimising Deformation Fields for Meshes}
To demonstrate the versatility of our approach, we apply it to meshes as well. We adapt the energy term and use surface samples and vertex positions for optimisation. For freestanding objects, we use curvature as the energy measure. For the room example in \cref{fig:seam_mesh}, the energy is set to $0$ for the floor and to $1$ for every other surface point. We otherwise follow the same principle as in \cref{sec:nerf}.


\subsection{Other Applications}\label{sec:editing}
While retargeting is our main contribution, our framework opens up multiple different editing opportunities that even extend the capabilities of seam carving \cite{classic_seam}.  The loss formulation is slightly adapted to implement the following exemplary operations, which are applicable to any domain and can serve as the basis for additional processing methods or user-driven applications. Further details can be found in Sec.~ D (Appendix). Examples can be found in \cref{fig:editing}. We want to remove the monotonicity loss $L_m$ and use equal input and output sizes for both the removal and the movement of an object.\\ 
\textbf{Removal} ~~ 
To remove a certain region, we employ an additional loss term that penalises any deformation target leading to that region. Points that were originally assigned to the target region will be distorted and assigned to neighbouring regions, resulting in the removal of the region in the output.\\
\textbf{Moving an Object} ~~ 
Although classic seam carving cannot move an object easily, as seams are assumed to cross the entire image, our formulation does not have this restriction. To move an object, we also enforce that the deformation towards the target feature only occurs from the desired new placement in the output. Any occurrence of that feature elsewhere is punished to avoid its repetition.

\begin{figure*}[t!]
    \includegraphics[height=0.135\linewidth]{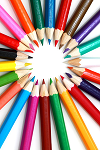}
    \includegraphics[height=0.135\linewidth]{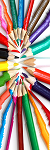}
    \includegraphics[height=0.135\linewidth]{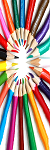}
    \includegraphics[height=0.135\linewidth]{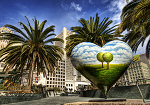}
    \includegraphics[height=0.135\linewidth]{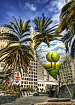}
    \includegraphics[height=0.135\linewidth]{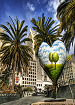}
    \includegraphics[height=0.135\linewidth]{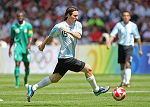}
    \includegraphics[height=0.135\linewidth]{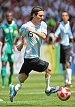}
    \includegraphics[height=0.135\linewidth]{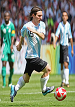}
    \caption{
        Selected results showing scenes where our approach benefits from a higher degree of flexibility. All images are arranged as input, seam carving, and our output. Images from \cite{retargetme}. 
    }\label{fig:2d_comp}
\end{figure*}

\begin{table*}[t]
\centering
\caption{Comparison of FID scores on the RetargetMe \cite{retargetme} dataset, for retargeting x (top) and y (bottom) axis to a smaller (left) and bigger (right) size. In mean, ours outperforms seam carving with $\textbf{46.67}$ to $52.57$ for x and $\textbf{53.57}$ to $65.48$ for y.}
\label{tab:retarget_only}
\begin{tabular}{|c|c|c|c|c|c||c|c|c|c|c|}
\hline
$x$ FID($\downarrow$)& 50\% & 60\% & 70\% & 80\% & 90\% & 110\% & 120\% & 130\% & 140\% & 150\% \\
\hline
Seam carving & 85.75 & 66.41 & 51.67 & 35.20 & 23.80 & 20.86 & 30.01 & 37.38 & 43.45 & \textbf{46.37} \\
\hline
Ours & \textbf{79.39} & \textbf{58.06} & \textbf{43.69} & \textbf{30.53} & \textbf{21.71} & \textbf{11.17} & \textbf{19.42} & \textbf{29.19} & \textbf{37.87} & 49.74 \\
\hline
\hline
$y$ FID($\downarrow$)& 50\% & 60\% & 70\% & 80\% & 90\% & 110\% & 120\% & 130\% & 140\% & 150\% \\
\hline
Seam carving & 116.70 & 84.33 & 60.59 & 40.86 & \textbf{24.92} & 23.04 & 33.58 & 40.82 & 48.49 & 52.59  \\
\hline
Ours & \textbf{89.44} & \textbf{67.77} & \textbf{49.62} & \textbf{35.59} & 25.42 & \textbf{13.55} & \textbf{21.02} & \textbf{27.55} & \textbf{38.20} & \textbf{46.40} \\
\hline
\end{tabular}
\end{table*}
\begin{table}[t]
    \centering
    \begin{minipage}{0.48\textwidth}
        \centering
    \caption{User study comparing our approach to seam carving (\textit{SC}) for images. $19$ participants were questioned for preferences of random examples from the RetargetMe dataset\cite{retargetme}, casting $~1200$ votes.}\label{table:userstudy_2d}
        \begin{tabular}{|l||c|c|c|}
    \hline
    Width & Ours & SC & Draw \\
    \hline
    50 $\%$ & \textbf{48.48\%} & 20.87\% & 30.67\%\\
    150 $\%$ & 33.8\% & \textbf{36.27\%} & 29.92\%\\
    \hline
    \hline
    Height & Ours & SC & Draw \\
    \hline
    50 $\%$ & \textbf{49.5\%} & 24.25\% & 26.25\%\\
    150 $\%$ & \textbf{45.84\%} & 20.0\% & 34.15\%\\
    \hline
    \hline
    Total & \textbf{44.57}\% & 25.1\% & 30.32\%\\
    \hline
    \end{tabular}
    \end{minipage}
    \begin{minipage}{0.48\textwidth}
        \centering
            \begin{tabular}{ccc}
                ~ &
                Colour &
                Complex \\
                Input &
                Gradient &
                Energy\\
                \includegraphics[height=0.30\linewidth]{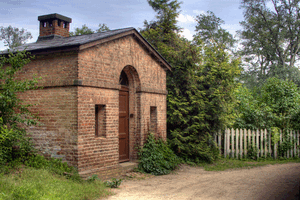}&
                \includegraphics[height=0.30\linewidth]{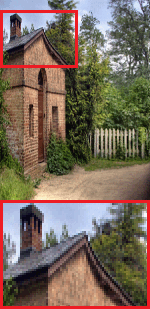}&
                \includegraphics[height=0.30\linewidth]{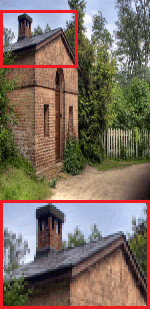}
            \end{tabular}
        \captionof{figure}{Example of using our method with the more complex energy term of \cite{srinivas2019full} instead of gradients, improving the results on an example.}\label{fig:other_energy}
    \end{minipage}
\end{table}
\begin{table}[t]
    \centering
    \caption{User study comparing retargeting a scene, then recording it (ours), to recording it and then retargetet it (video seam carving \cite{video_seam}, \textit{VSC}), using our provided dataset. $21$ participants were asked to indicated which solution they prefer, casting $252$ votes in total.}\label{table:userstudy_3d}
    \begin{tabular}{|c||c|c|c||c|c|c||c|c|c||c|c|c|}
        \multicolumn{1}{c}{} & \multicolumn{3}{c}{X Axis to 50\%} & \multicolumn{3}{c}{X Axis to 150\%} & \multicolumn{3}{c}{Y Axis to 50\%} & \multicolumn{3}{c}{Y Axis to 150\%} \\
        \hline
        Scene  & Ours & VSC & Draw & Ours & VSC & Draw & Ours & VSC & Draw & Ours & VSC & Draw \\
        \hline
        Beach & \textbf{17} & 3 & 1 & \textbf{20} & 1 & 0 & \textbf{19} & 1 & 1 & \textbf{19} & 1 & 1 \\
        Farm & \textbf{21} & 0 & 0 & \textbf{21} & 0 & 0 & \textbf{16} & 3 & 2 & \textbf{21} & 0 & 0 \\
        Harb. & \textbf{21} & 0 & 0 & \textbf{21} & 0 & 0 & \textbf{21} & 0 & 0 & \textbf{21} & 0 & 0 \\
        \hline
        \multicolumn{7}{c}{}
    \end{tabular}
\end{table}
\section{Evaluation}
Retargeting not only lacks comparable methods for other visual data such as meshes, but its quality is also difficult to measure. In 2D, we compare our results to seam carving~\cite{classic_seam} on images using the \emph{RetargetMe}~\cite{retargetme} benchmark. In 3D, we compare our results on NeRFs with video-seam-carved recordings of the same scene, because video seam carving~\cite{video_seam} is the closest method to 3D retargeting. We always produce a short, $~3$ second clip at $30$ frames per second. To ensure a fair comparison with the method, we only tilt the camera down instead of rotating it around the object. Even for benevolent camera movements, it produces visible artefacts, as seen in Sec.~ F (Appendix). Exemplary, we show that our approach is compatible with approaches that propose different saliency maps as visualised in \cref{fig:taxonomy}. For that purpose, we compare our method to other more recent image retargeting approaches visually (\cref{fig:other_energy}) and in a user study (\cref{table:userstudy_ours_vs_theirs}). Additional results can be found in Sec.~ G (Appendix), including an explicit failure case.
\subsection{Comparisons}\label{sec:eval_qualitative}\label{sec:eval_quanitative}
\subsubsection{Images}
In accordance with recent publications in the generative model and image editing domain \cite{gu2023matryoshka, peebles2023scalable, kawar2023imagic}, we extensively compare our results with the original seam carving results using FID \cite{heusel2017gans} in \cref{tab:retarget_only} for images. We also evaluate our approach in a user study. The participants were instructed to classify their preference for a retargeted version of a randomly selected image based on a randomly selected deformation axis, always choosing between our and the seam carved\cite{classic_seam} version. We used the $80$ images of the RetargetMe~\cite{retargetme} for 2D, producing $360$ variations per approach in \cref{table:userstudy_2d}. This shows how our approach can be a more powerful backbone for seam carving.\\
\begin{wrapfigure}{l}{0.55\textwidth} 
  \centering
  \captionof{table}{User study measuring preference, pairwise comparing to our approach using salience maps of \cite{srinivas2019full}.}\label{table:userstudy_ours_vs_theirs}
  \begin{tabular}{|l||c|c|c|}
      \hline
      Approach & Ours & Theirs & Draw \\
      \hline
            CarvingNet\cite{SongLL19} & \textbf{45.16\%} & 41.93\% & 12.91\%\\
            RL\cite{rl_seamcarve} & \textbf{46.85\%} & 32.43\% & 20.72\%\\
            Cycle-IR\cite{tan2019cycle} & \textbf{40.3\%} & 34.46\% & 25.24\%\\
      \hline
  \end{tabular}
\end{wrapfigure}
We further compare against other state-of-the-art retargeting approaches in a user study in \cref{table:userstudy_ours_vs_theirs}: We show that our method can compete both against a seam carving variation with better saliency from deep learning~\cite{SongLL19} while also beating state-of-the-art approaches that retarget the image directly~\cite{tan2019cycle, rl_seamcarve}. Note that we only compare to readily available images from other approach, as we often could not reproduce them or run the code. For this comparison, we use the saliency method FullGrad~\cite{srinivas2019full} that aggregates layer-wise bias-gradients from across ResNet~\cite{resnet}. This aggregated gradient map acts as our saliency map, as it measures how useful the feature is to the neural network, hence we use it as a measure for how semantically important a feature is. This also shows that our approach can work with different saliency maps. Details and examples can be found in Sec.~ E (Appendix). Additionally, we provide a visual ablation that shows the importance of each loss component in \cref{fig:losses}. 
\subsubsection{3D Scenes}
When assessing the quality of retargeting 3D scenes provided as NeRFs, we compared our results with what we consider to be the most similar approach, video seam carving~\cite{video_seam}. In our provided GTA V NeRF retargeting dataset, we resize each axis by $50\%$ and $150\%$. The results can be found in \cref{table:userstudy_3d}. We provide additional results and frames of the resulting sequences from our approach for 3D scenes compared to video seam carving in Sec.~ F (Appendix).%

\begin{figure}[t]
	\centering
	{
		\begin{tabular}{cc}
                \includegraphics[trim={0cm 2.5cm 0 0},clip,width=0.26\linewidth]{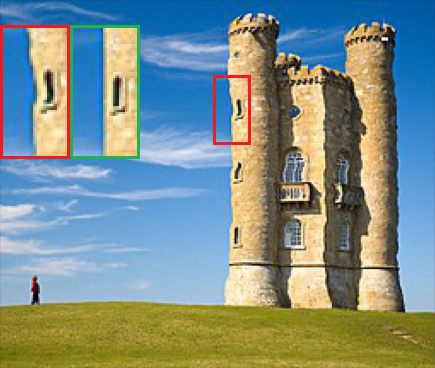} 
                \includegraphics[trim={0cm 2.5cm 0 0},clip,width=0.26\linewidth]{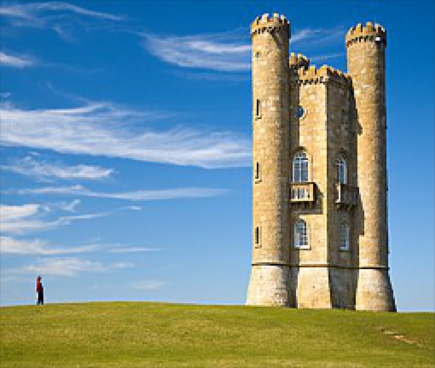}
                \includegraphics[trim={0 0 1cm 0cm},clip,width=0.20\linewidth]{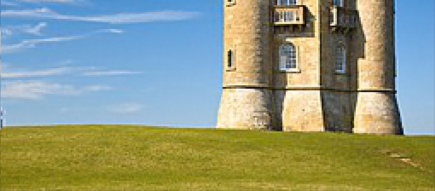}
                \includegraphics[trim={0 0 1cm 0cm},clip,width=0.20\linewidth]{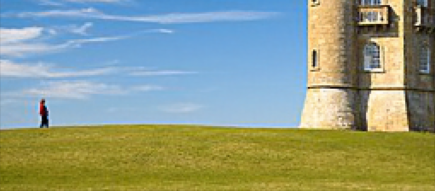}
		\end{tabular}
	}
    \caption{Visual ablation for retargeting to $80\%$ width on an image\cite{tower}, from left to right: 
    Wobbly contours from shearing (no $L_s$), uniform stretch on the whole image (no $L_e$), missing features \eg the person (no $L_m$), content outside the image (no $L_b$).}
    \label{fig:losses}
\end{figure}


\subsection{Discussion and Limitations} 
With our deformation fields as backbone, but the same energy term, our approach outperforms seam carving on both a qualitative and quantitative level. Our flexibility in the energy formulation allows for more possibilities, and the replacement of the gradient-based energy formulation with a better saliency map improves our approach even more. This shows that our method is a flexible backbone that can be used with various successors to seam carving, which provide more advanced saliency maps. Our continuous formulation allows for retargeting in various domains, including 3D scenes, and highlights the advantages of a geometry-aware approach. Applying our approach with only colour gradients can produce flawed outputs, just as those of seam carving, see Fig.~ 15 (Appendix). As we optimise a neural field for \emph{each individual input}, we trade computation time for not needing a dataset to train and for a global optimisation. Our deformation optimisation takes approximately one minute on consumer hardware for good results versus about half a second for seam carving. Our strength, however, lies in 3D. Video seam carving on a 600 by 400 pixel video can take several hours, whereas our approach produces the final retargeted 3D scene in under 15 minutes. For Future Work, investigating invertible neural networks to implicitly learn an inverse deformation for the 3D case \cite{invertible} seems promising, as would a deep learned approach to evaluating a deformation for optimisation and more research into good measures for retargeting quality.
\section{Conclusion}
We present an approach that generalises the principle underlying seam carving to create image variants by manipulating low content areas. By training neural deformation fields that avoid applying distortion to important parts of an input, our approach is applicable to different forms of visual data retargeting. Our formulation is largely domain invariant and only needs minor alterations to work on different types of visual data, as demonstrated for neural radiance fields and meshes for the first time. We demonstrate compatibility to other energy formulations of the seam carving family, being a more general way of optimising image retargeting. Our approach is the only one that natively works on 3D scenes.
%
%
\bibliographystyle{splncs04}

\clearpage
\setcounter{page}{1}
\setcounter{section}{0}
\renewcommand{\thesection}{\Alph{section}}


\section{Architecture}\label{app:architecture}
\begin{figure}[h]
\centering
\includegraphics[width=0.6\textwidth]{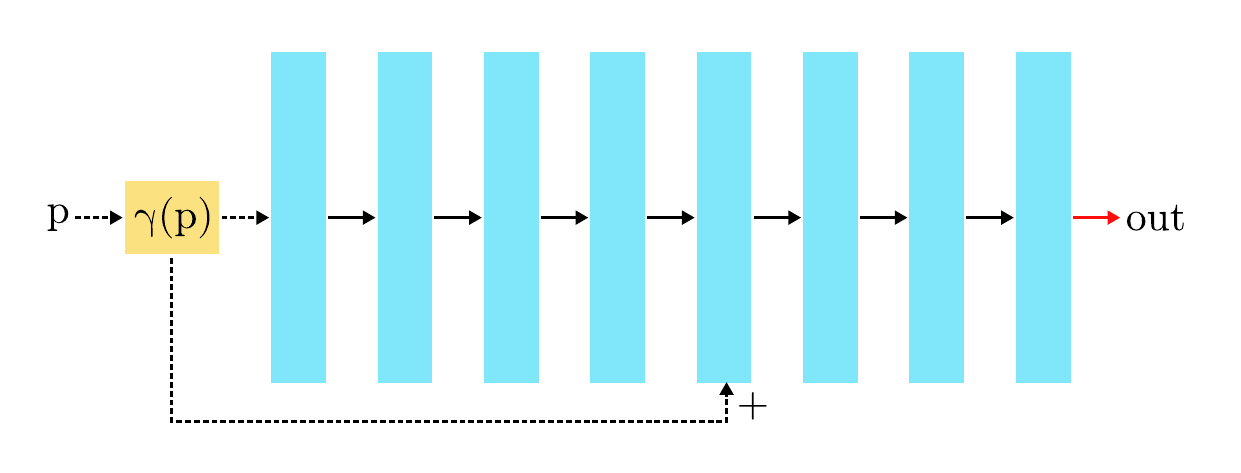}
\caption{The MLP we apply for learning the neural image field, the deformation field(s), energy network and cumulative energy network. We use positional encoding (yellow), then apply linear layers (cyan) with a residual connection, LeakyReLu (black arrow), and an output depending on context (red, with sigmoid for images, otherwise LeakyReLu).}\label{fig:architecture_imgs}
\end{figure}
We use simple components for our architecture, a basic MLP with residual connection being the core of all our used networks. We use the Adam optimiser\cite{adam} in all instances, without weight or learning rate decay, and a default learning rate of $0.001$ if not specified otherwise.
\subsection{Architecture for Images}
For the neural field holding the images as described in \cref{sec:images}, we apply the network depicted in \cref{fig:architecture_imgs}, a simple MLP with 192 channels, a residual connection, and positional encoding that outputs $3$ scalars for the RGB value. For the deformation and the cumulative gradient network, we use the same network architecture with 64 channels and a single output scalar value. We use a learning rate of $0.0001$ for image expansion. We train the network learning the image itself and the network learning the cumulative gradient for $250$ epochs with $100$ iterations each. We initialise the deformation network for $50$ epochs with a uniform transformation that we then train on our loss defined in \cref{app:losses}. Our slim architecture allows optimisation of all the image data in one single batch. We generally observe almost the same quality for significantly fewer epochs.
\subsection{Architecture for Neural Radiance Fields}
We use a pre-trained, custom iNGP \cite{muller2022instant} as our learned 3D scene and apply the same setup as for images. However, both vanilla NeRF~\cite{mildenhall2020nerf} and mip-NeRF~\cite{barron2021mip} worked just as well in our early tries. We train the initial deformation, energy, and cumulative energy for 5000 iterations each, always using $10000$ samples. We optimise the deformation field for $10000$ random points from the surface and $10000$ random points from the boundary for 50 epochs with 100 iterations each, and always choose the network parameters from the epoch with the best average loss.


\section{Exact Loss Formulations}\label{app:losses}
\subsection{Loss Formulation for Images}\label{app:losses_image}
Due to the lightweight network we use for our deformation field, we can apply our loss terms directly to all points $p$ in the output domain (\ie the shrunken image), training one single batch for an image. 

For $L_e$, we define
\begin{equation}\label{app:exact_loss_img}
\begin{aligned}
    L_e = \frac{|\Sigma(p) - \Sigma(p_\varepsilon)| \cdot |D(p) - D(p_\varepsilon)|}{\varepsilon},
\end{aligned}
\end{equation}
where $\Sigma$ is the MLP for cumulative energy and $p_\varepsilon$ are all pixels $p$ moved by one pixel in direction $v$. For shearing, we define
\begin{equation}\label{app:exact_loss_shear}
\begin{aligned}
    L_s &= \frac{E(p) \cdot |D(p) - D(p_\varepsilon^{\perp})|}
    {\varepsilon},
\end{aligned}
\end{equation}
where $E$ is the energy net and $p_\varepsilon^{\perp}$ are all points $p$ moved by one pixel in the direction orthogonal to $v$.
For the boundary, we build our loss around the pixels on the one end of the image, $p_l$, and the other, $p_r$, regularising their deformation to be $0$ at the one, and $1.0 - \alpha$ at the other end:
\begin{equation}\label{app:exact_loss_b}
    L_b = mean(|D(p_l)|) + mean(|D(p_l) - (1 - \alpha)|)
\end{equation}
For monotonicity in images, we use:
\begin{equation}\label{app:exact_loss_m}
    L_m = \frac{mean(max(0, D(p) - D(p_\varepsilon)))}{\varepsilon}
\end{equation}
As weights for our loss terms, we use $\lambda_{e} = 10000,~\lambda_{s} = 250~\lambda_{b} = \lambda_{m} = 10000$.
For the cumulative energy net, we accumulate energy over all pixels, then train the MLP to minimise mean squared error between input (2D position) and output (cumulative energy for that position) for 10000 iterations, always for all pixels.

\subsection{Loss Formulation for Neural Radiance Fields}\label{app:losses_nerf}
We first define a mixture of surface points and uniform random points $p'$ in source space, on which we run our optimisation on, then find those points $p$ in target space that are deformed to become $p'$:\\
\begin{equation}\label{eq:u}
    p = U(p')
\end{equation}
For $L_e$, we define
\begin{equation}\label{app:nerf_exact_loss_img}
\begin{aligned}
    L_e = \frac{mean(|\Sigma(p) - \Sigma(p_\varepsilon)| \cdot |D(p) - D(p_\varepsilon)|)}{\varepsilon},
\end{aligned}
\end{equation}
where $\Sigma$ is the MLP for cumulative energy and $p_\varepsilon$ are our points $p$ moved by an offset in direction $v$. For shearing, we define
\begin{equation}\label{app:nerf_exact_loss_shear}
\begin{aligned}
    L_s = \frac{E(p) \cdot |D(p) - D(p_{\varepsilon^1}^{\perp})|}
    {\varepsilon}+ \frac{E(p) \cdot |D(p) - D(p_{\varepsilon^2}^{\perp})|}
    {\varepsilon},
\end{aligned}
\end{equation}
where $E$ is the energy net and $p_{\varepsilon^1}^{\perp}$, $p_{\varepsilon^2}^{\perp}$ are all points $p$ moved by an offset in the directions orthogonal to $v$.
For the boundary, we build our loss around the pixels on the one end of the image, $p_l$, and the other, $p_r$, regularising their deformation to be $0$ and $1.0 - \alpha$ respectively:
\begin{equation}\label{app:nerf_exact_loss_b}
    L_b = mean(|D(p_l)|) + mean(|D(p_l) - (1 - \alpha)|)
\end{equation}
For monotonicity, we use:
\begin{equation}\label{app:nerf_exact_loss_m}
    L_m = \frac{mean(max(0, D(p) - D(p_\varepsilon)))}{\varepsilon}
\end{equation}
As weights for our loss terms, we use $\lambda_{e} = 10,~\lambda_{s} = 0.1~\lambda_{b} = 100, \lambda_{m} = 1$. After every iteration, we also update $U$ with all points $p$, \ie minimise $(D(U(p)) - p)^2$ for one step.
For the cumulative energy net, we accumulate energy by shooting random rays along the deformation direction with $100$ uniformly distributed samples. We gather and accumulate energy from $E$, then train the cumulative energy MLP to minimise mean squared error between input (3D position) and output (cumulative energy for that position on that ray) for $10000$ iterations, always for $10000$ points.



\section{Expansion}\label{app:expansion}
Expansion of visual data follows the same principle as in seam carving\cite{classic_seam}: Instead of compressing (in seam carving: removing) low energy parts, we stretch (in seam carving: double) them. Examples can be seen in \cref{app:extra_results}. To expand data, \eg to increase its width by $150\%$, we can use the exact same losses as for retargeting to a smaller size, but with one additional loss term and a minor modification.\\
We require a limit on the change in the deformation to avoid only duplicating a single seam. It should punish any change in the deformation that is too large, hence would repeat the same part of the image over and over again. Seam carving prevents this for image expansion by only doubling a seam once. As a new loss term, this gives us:
\begin{equation}
    L_{cap} = mean\left(max\left(0, \frac{(p - p_{\varepsilon})}{\varepsilon} - 1\right)\right)
\end{equation}
This formulation prevents an increase of more than double the input. In addition, we modify $L_e$ to not punish any information content skipped between two points (as we no longer skip over anything), but instead simply punish the energy directly, \ie replacing $\hat{E}$ from \cref{eq:energy_between} with $E$ applied to only one point.


\section{Editing}\label{app:editing}
For the editing procedures describe in \cref{sec:editing}, we can perform the same concept for every domain. For this, we always retain the same image size.
\paragraph{Removal} To remove an object, we punish every point that maps to the content that we want to remove. Meaning, we take all points of the input, deform them, and look up if the deformed point lies within the marked area. For continuity, we do this by training another network $net_{mask}$ to learn a mask, \ie $0$ for regions outside, $1$ for regions inside the area to remove. We then simply add $mean(net_{mask}(p))$ for all points $p$ to the loss function, punishing any deformation that maps inside the target area. To allow both sides of the removed content to fill in the space, we disable the monotonicity loss $L_m$. The remaining loss terms then keep the rest of the image in place.\\
\paragraph{Moving an object} To move an object, we first remove the monotonicity loss $L_m$ to allow more complex deformations. As this disables protection against repetition, we actively punish our target object occurring anywhere else. For that, we utilise a network $net_{mask}$ containing a binary mask of the object to move, then add $net_{mask}(p)$ for all points $p$ in the output to the loss function except for the target coordinates of the object. We then add a term to the loss that enforces the exact offset value that would place the target at the right location for all parts of it. To avoid any other unwanted edits and make the optimisation more stable, we add one last loss term scaled by $0.01$ that punishes the absolute mean of deformation, \ie punishes applying any deformation at all. While this avoids unnecessary deformations, it is negligible compared to the additional loss caused by only the part relevant for the deformation.\\
To move an object along multiple axis, we decompose the movement to the target position into two axes, then apply our approach for each direction.

%
\newpage
\section{Comparison to Other Approaches}\label{app:comp_other_approaches_new}
For Carving-Net~\cite{SongLL19}, we only could access $4$ examples, whereas for Cycle-IR~\cite{rl_seamcarve}, we only found $29$ examples in their repository. Our randomised user study collected about $500$ votes from $15$ different users. Resulting images can be seen in \cref{table:ours_vs_them_images}. Carving-Net~\cite{SongLL19} produces a saliency map with deep learning, then applies seam carving. Note that for Carving-Net, the original images were resized to a different aspect ratio (squared) and resolution while reducing to only $75\%$ and $50\%$. The original paper stated, incorrectly, that the images were reduced to $25\%$ of the original size. For that reason, the images look different in the table. We used their original images as baseline for us and applied the same retargeting size to keep the comparison fair.

\begin{table}
    \begin{center}
    \captionof{figure}{Examples of images adapted with different approaches. Carving-Net results in $50\%$/$75\%$/$50\%$ of the original width from an initially cropped image, all other images show reduction to $50\%$ of the original width. For the comparison with Carving-Net, we used the same input image and same target size as they did. The image marked by ``n/a'' was not available in the repository of Self-Play RL.}\label{table:ours_vs_them_images}
    \begin{tabular}{cccc}
        Carving-Net\cite{SongLL19} & Self-Play RL\cite{rl_seamcarve} & Cycle-IR\cite{tan2019cycle} & Ours\cite{video_seam}\\
        \includegraphics[height=0.2\linewidth]{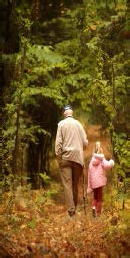}&
        \includegraphics[height=0.2\linewidth]{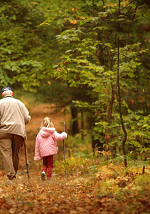}&
        \includegraphics[height=0.2\linewidth]{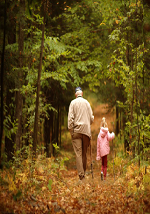}&
        \includegraphics[height=0.2\linewidth]{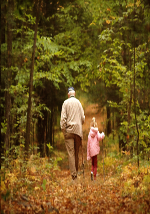}\\
        \includegraphics[height=0.2\linewidth]{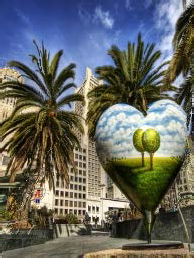}&
        n/a&
        \includegraphics[height=0.2\linewidth]{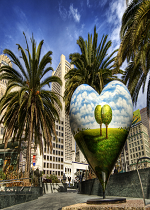}&
        \includegraphics[height=0.2\linewidth]{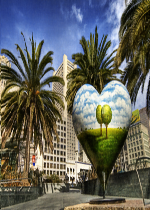}\\
        \includegraphics[height=0.2\linewidth]{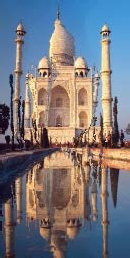}&
        \includegraphics[height=0.2\linewidth]{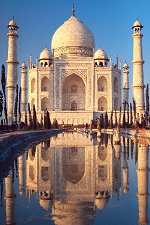}&
        \includegraphics[height=0.2\linewidth]{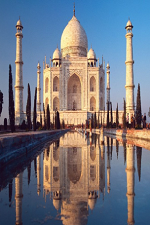}&
        \includegraphics[height=0.2\linewidth]{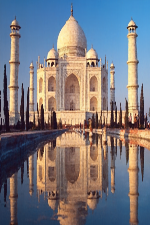}\\
    \end{tabular}
    \end{center}
\end{table}

\newpage
\section{Extended Results of Retargeting 3D scenes}\label{app:video_examples}
Example frames of retargeting 3D scenes can be seen in \cref{table:qualitative_vsc}. The figure compares the original images of a camera trajectory with the same camera trajectory on a deformed scene with our approach and with video seam carving\cite{video_seam} applied to the original scene recording. Note that the artefacts of video seam carving become highly visible from frame to frame. They result from video space seams rather than true geometry seams and from the local optimisation nature of video seam carving, that selects seams one-by-one to remove them in discrete seam carving approaches. While voxelised approaches of seam carving could be used, \eg with a similar graph cut algorithm like in video seam carving, our attempt realising that greatly suffered from high computational cost and voxelisation artefacts. These were the reason for us to switch from our voxelising to our current, continuous formulation that is also optimised globally. We also show quantitative results for 3D scenes in \cref{tab:fkthem} using FID.  These FID scores do not account for the strongly visible temporal artefacts between the frames from video seam carving. While we observe the results to align with our perception, we also validate this on a qualitative level through user studies in \cref{sec:eval_qualitative}.

\begin{table*}
    \begin{center}
    \caption{Quantitative results, measuring for deformation fields to retarget NeRFs versus video seam carving \cite{video_seam} on a video of the same camera trajectory. Note that temporal artefacts are not considered by FID.}
    \label{tab:fkthem}
    \begin{tabular}{|l||c|c|c|c|}
        \hline
        Scene FID($\downarrow$) & Beach & Farm & Harbour \\
        \hline
        X to 50\%, ours & \textbf{68.86} & \textbf{109.39} & \textbf{138.60} \\
        X to 50\%, VSC & 157.16 & 171.31 & 154.19 \\
        \hline
        Y to 50\%, ours & \textbf{79.70} & \textbf{147.46} & 181.97 \\
        Y to 50\%, VSC & 102.26 & 231.63 & \textbf{180.46} \\
        \hline
    \end{tabular}
    \end{center}
\end{table*}
\begin{table*}[ht]
    \begin{center}
    \begin{tabular}{ccccc}
Input & Ours & Ours, later time & VSC & VSC, later time\cite{video_seam}\\
\includegraphics[height=0.2\linewidth]{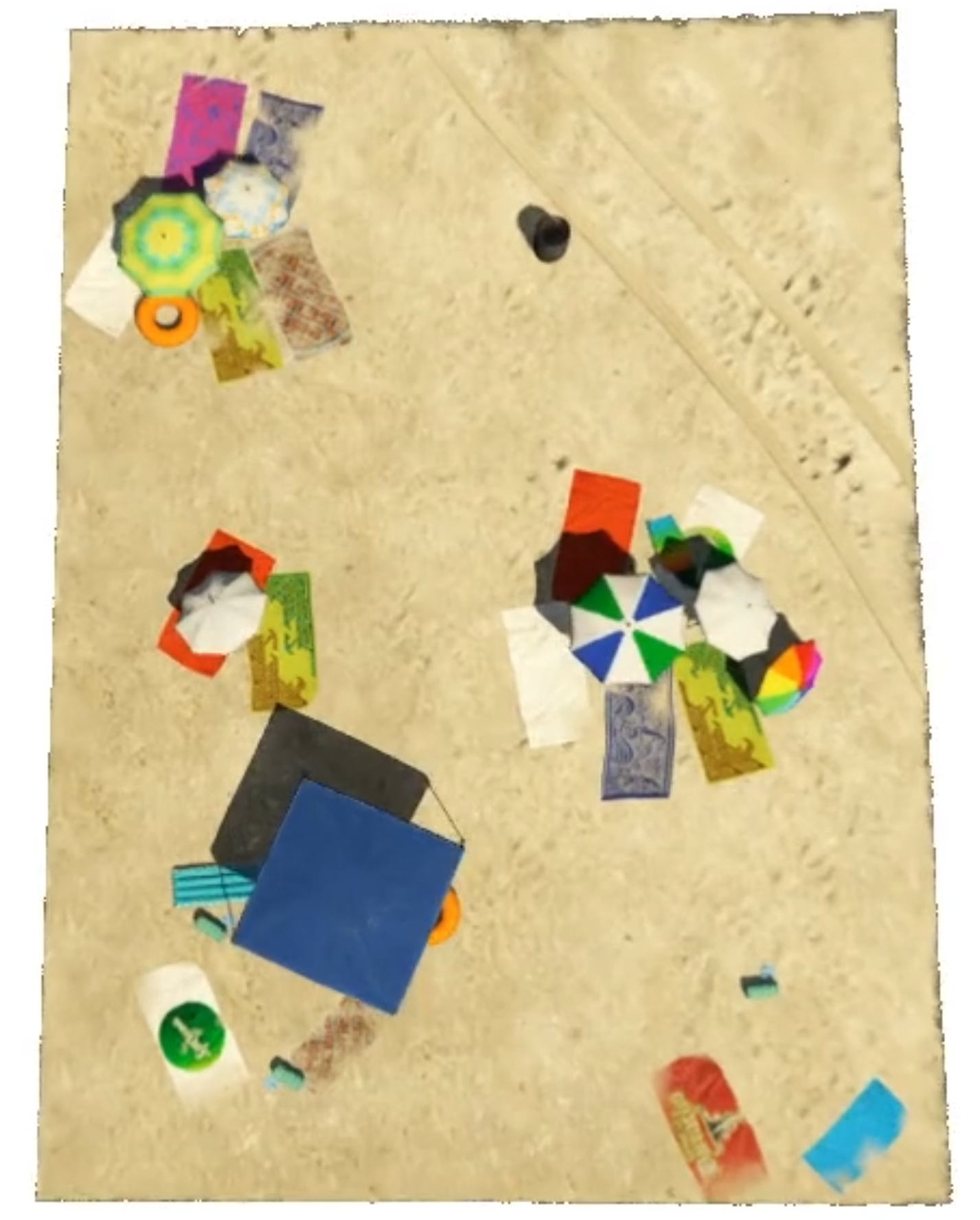}&
\includegraphics[height=0.2\linewidth]{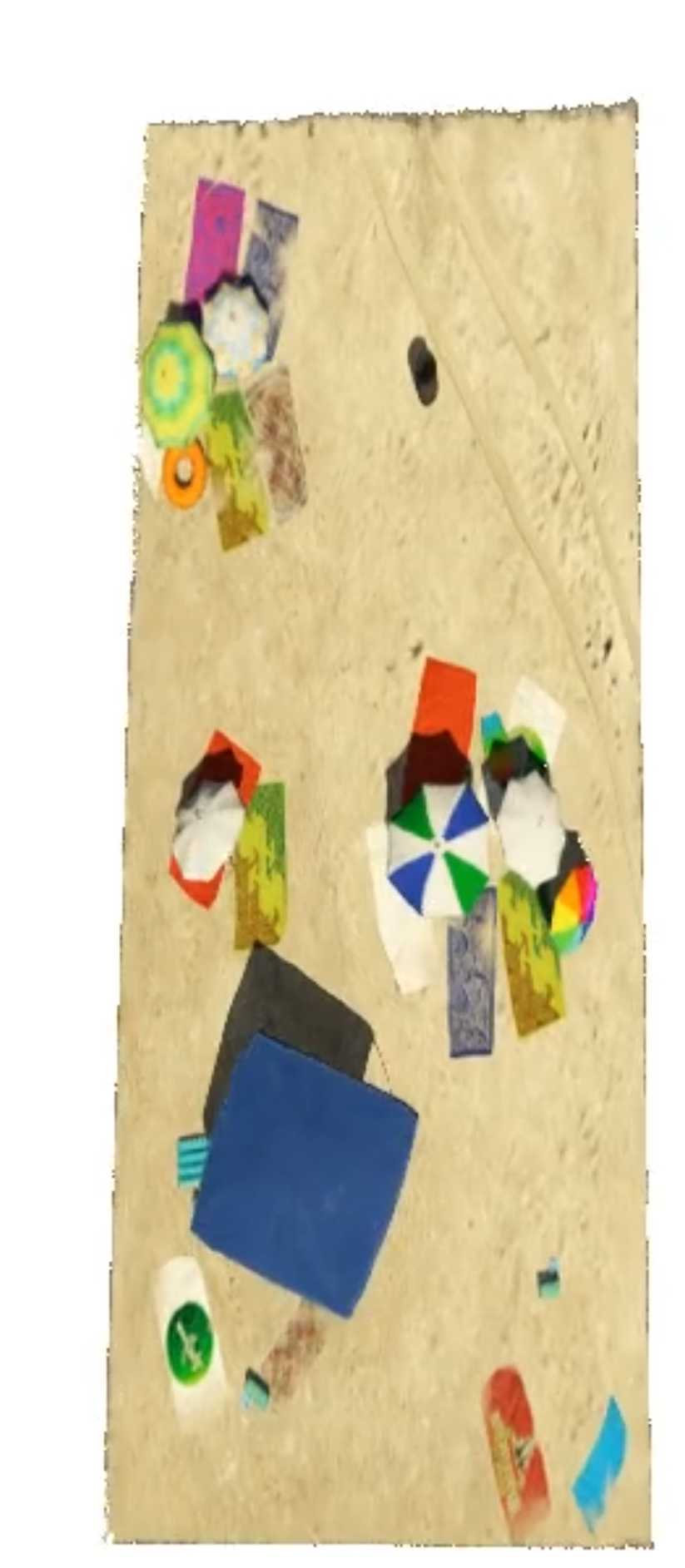}&
\includegraphics[height=0.2\linewidth]{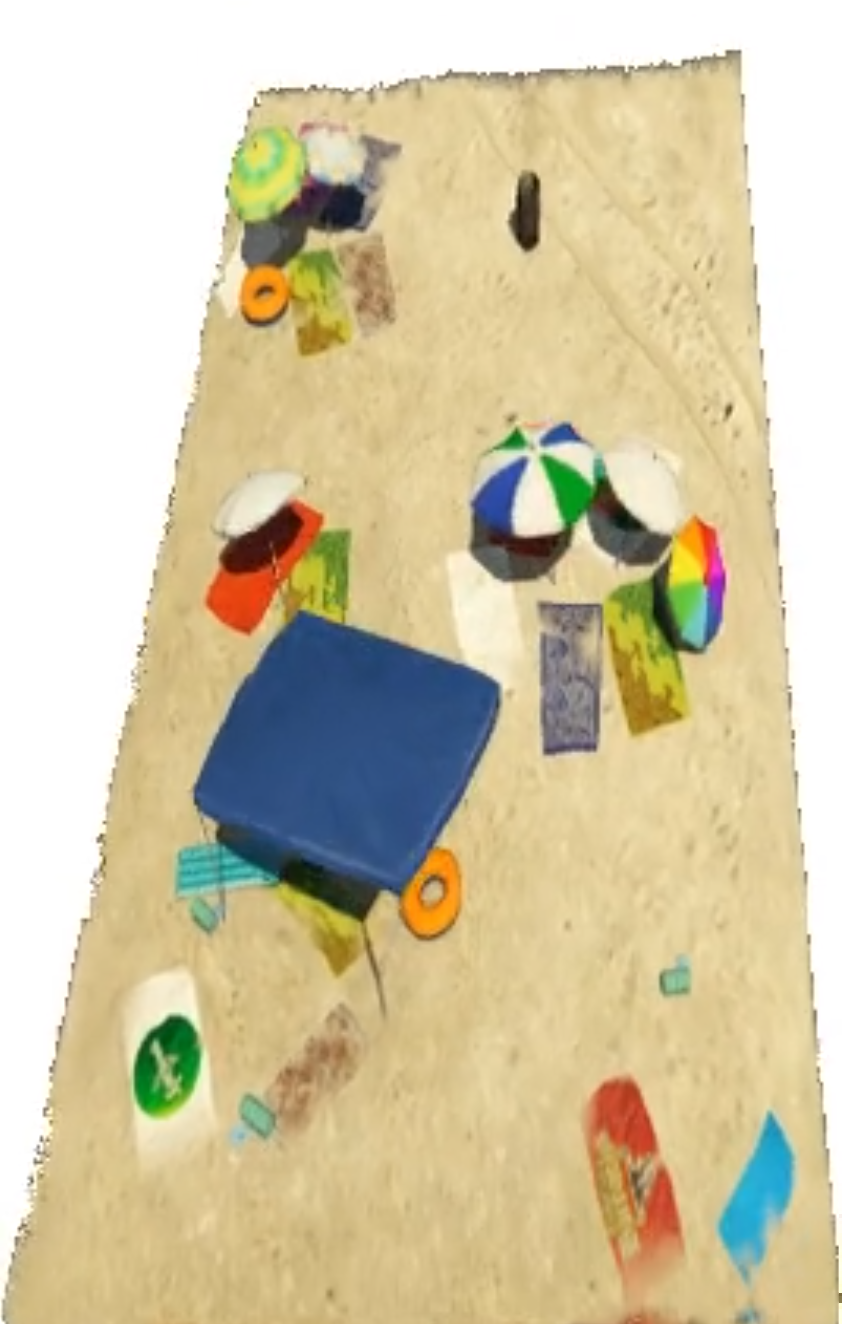}&
\includegraphics[height=0.2\linewidth]{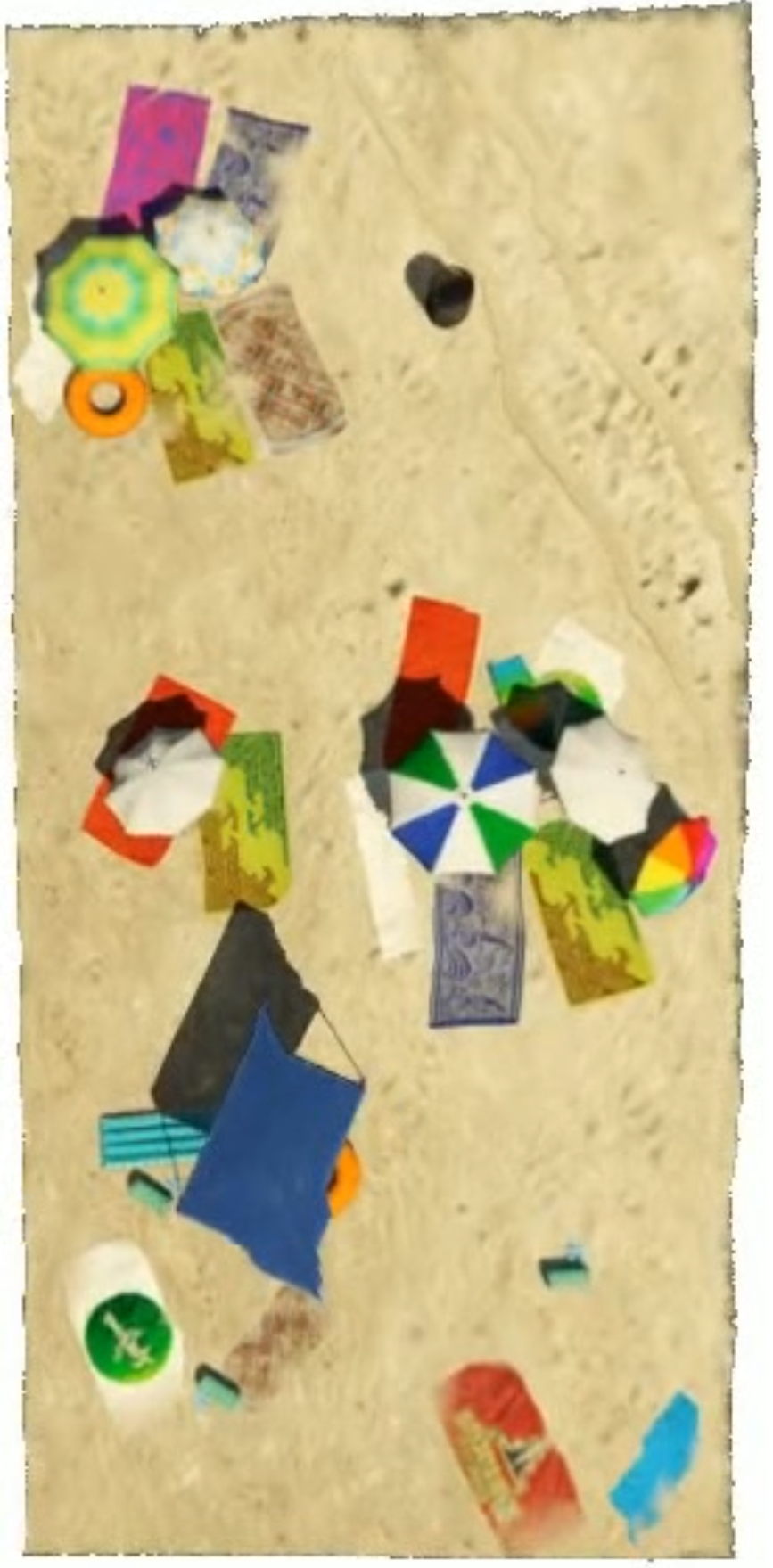}&
\includegraphics[height=0.2\linewidth]{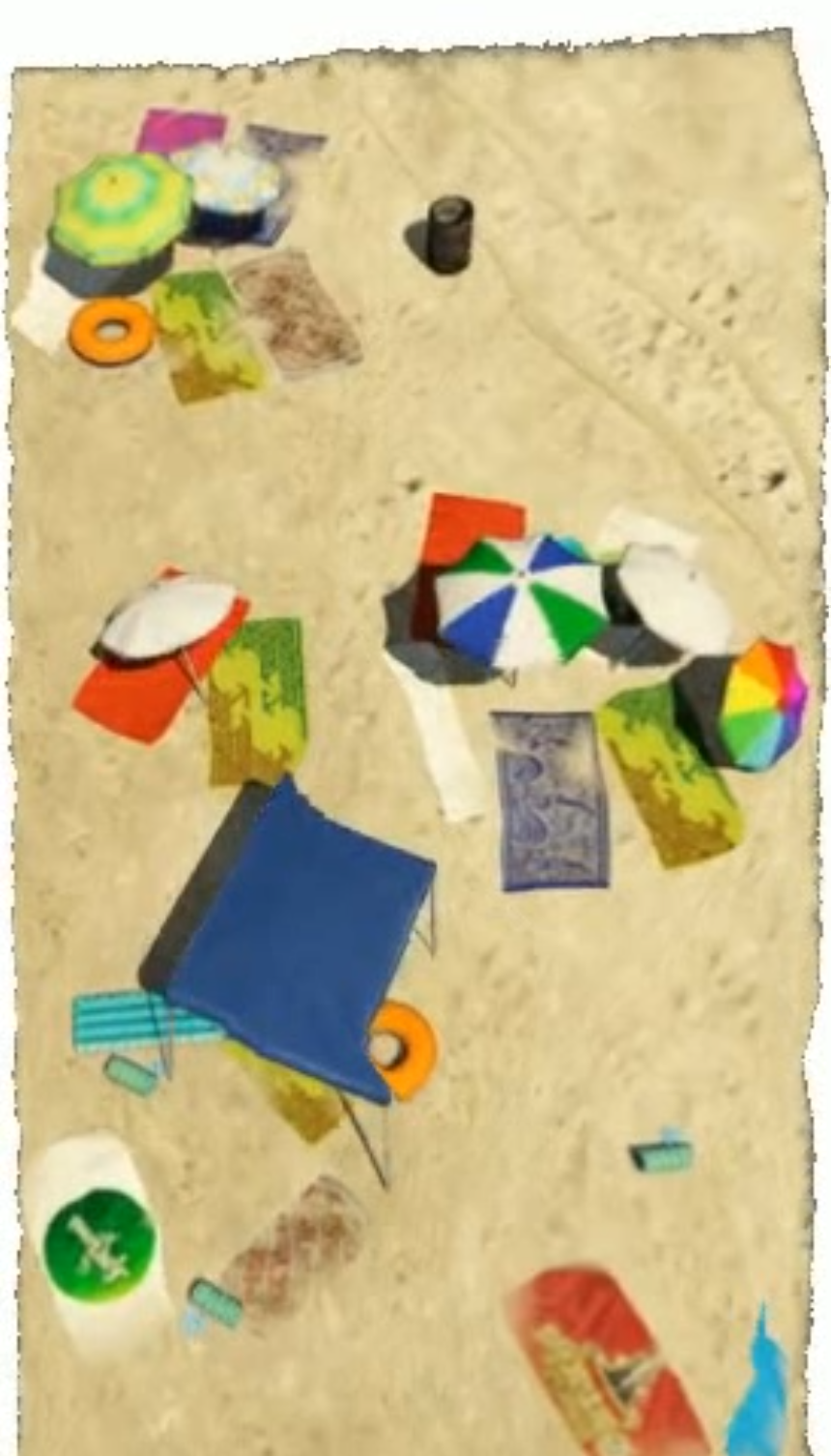}\\
\includegraphics[height=0.2\linewidth]{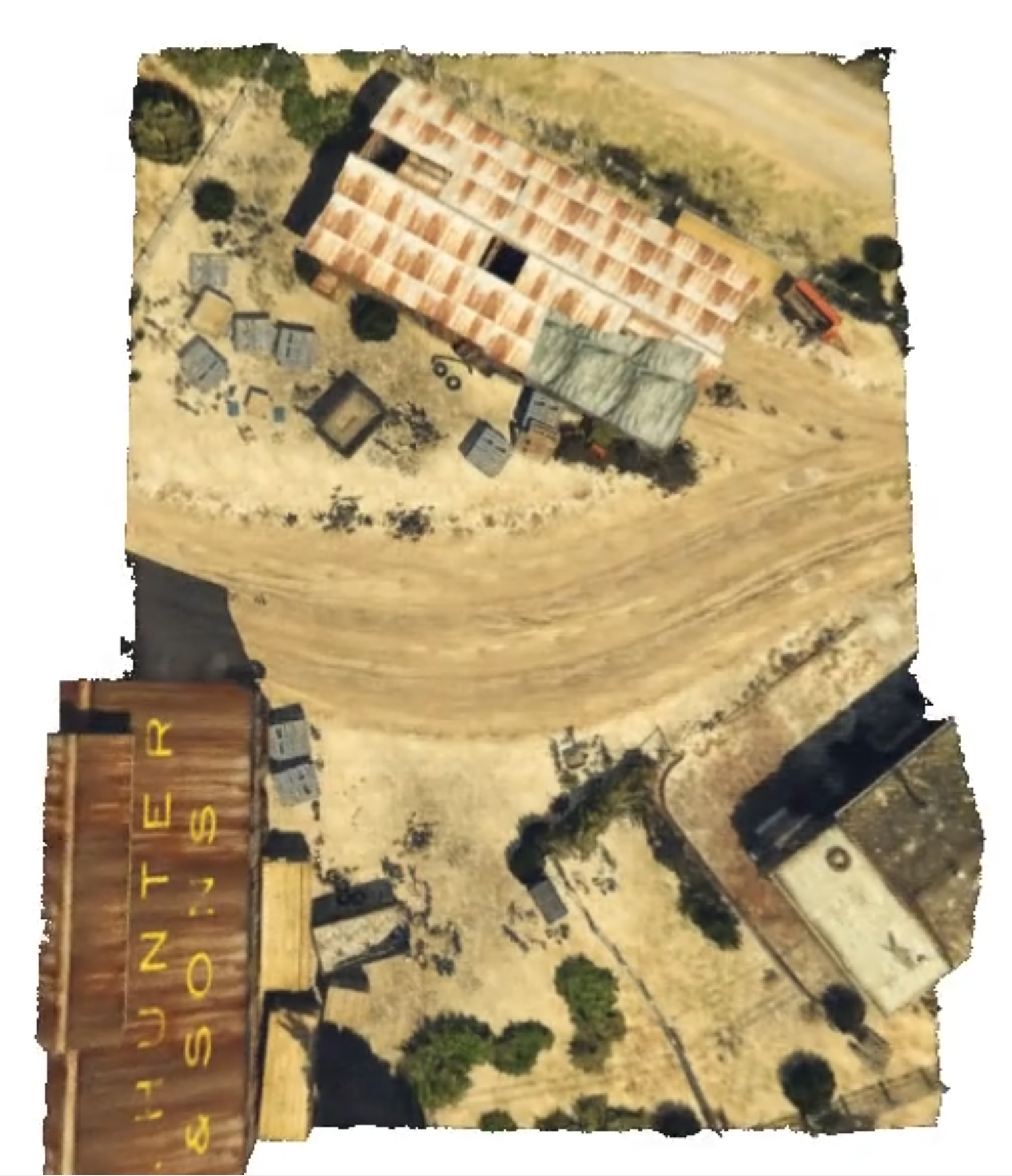}&
\includegraphics[height=0.2\linewidth]{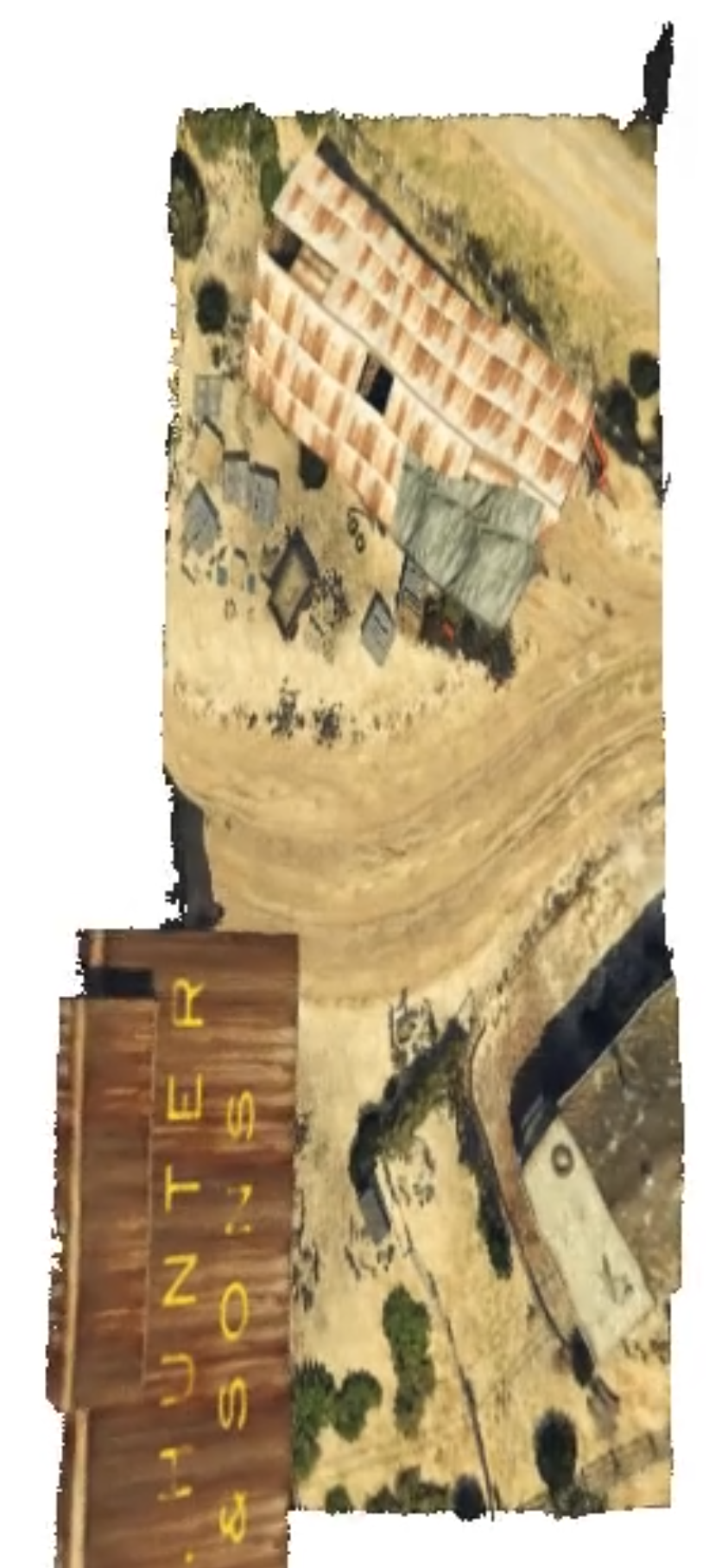}&
\includegraphics[height=0.2\linewidth]{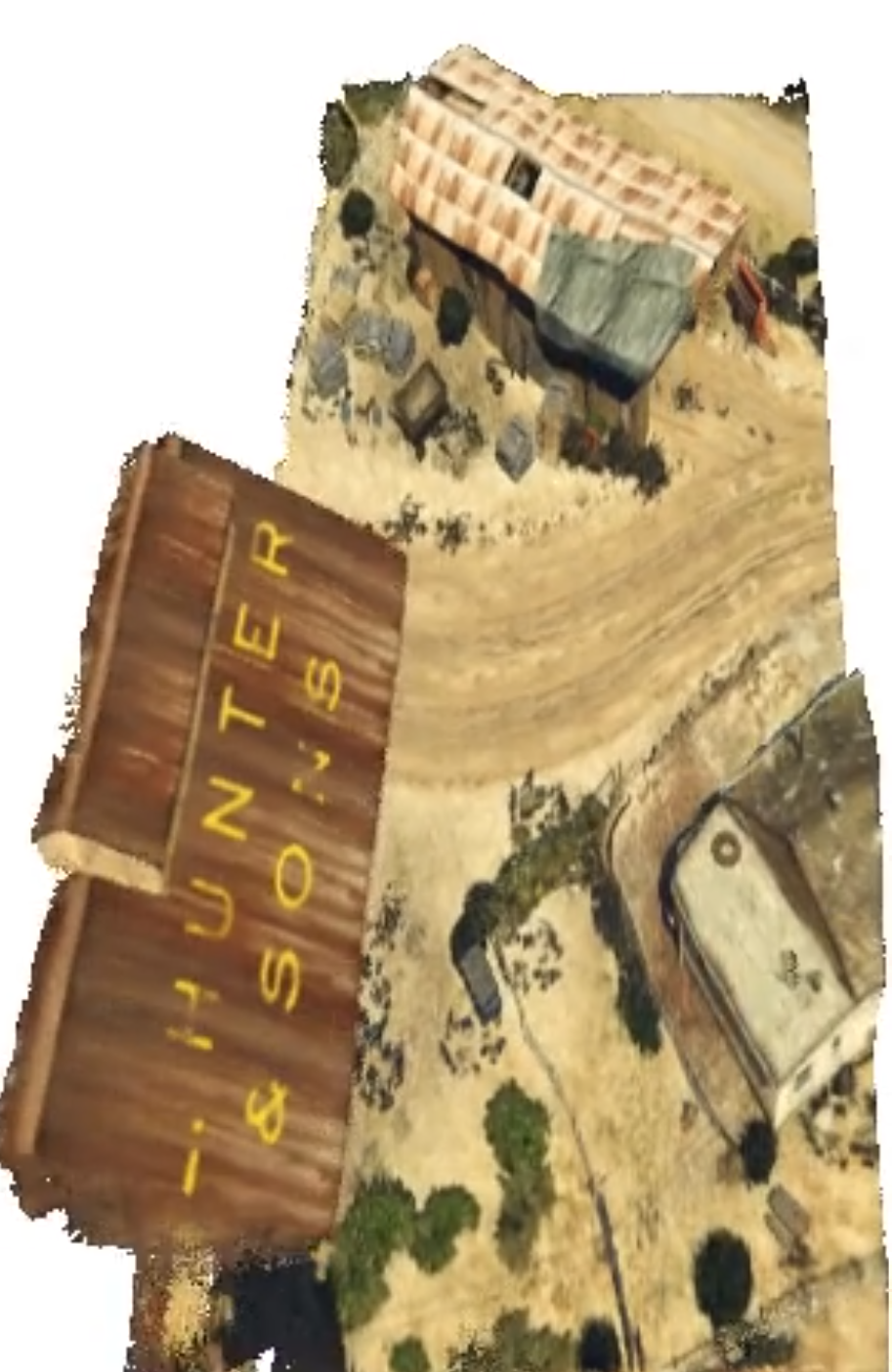}&
\includegraphics[height=0.2\linewidth]{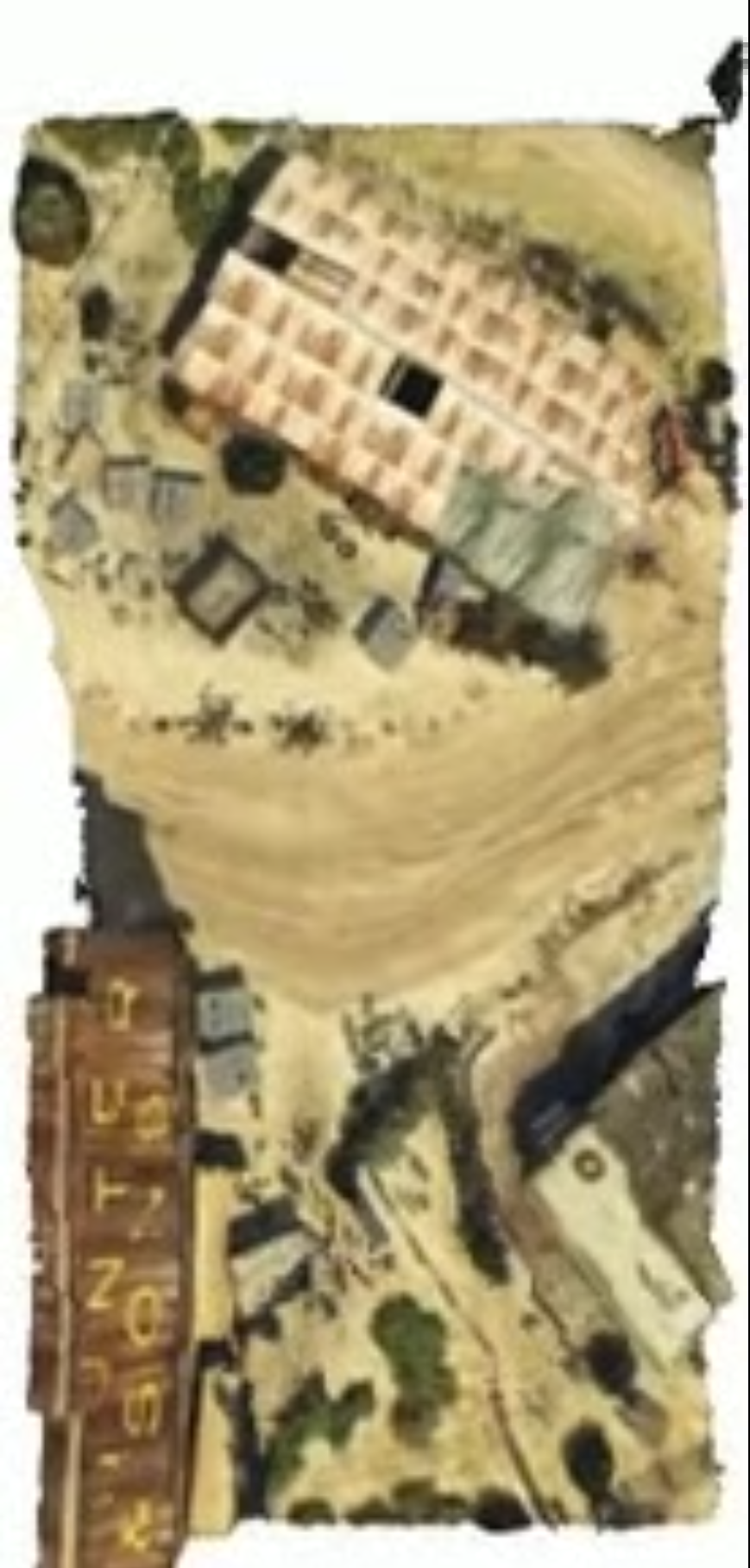}&
\includegraphics[height=0.2\linewidth]{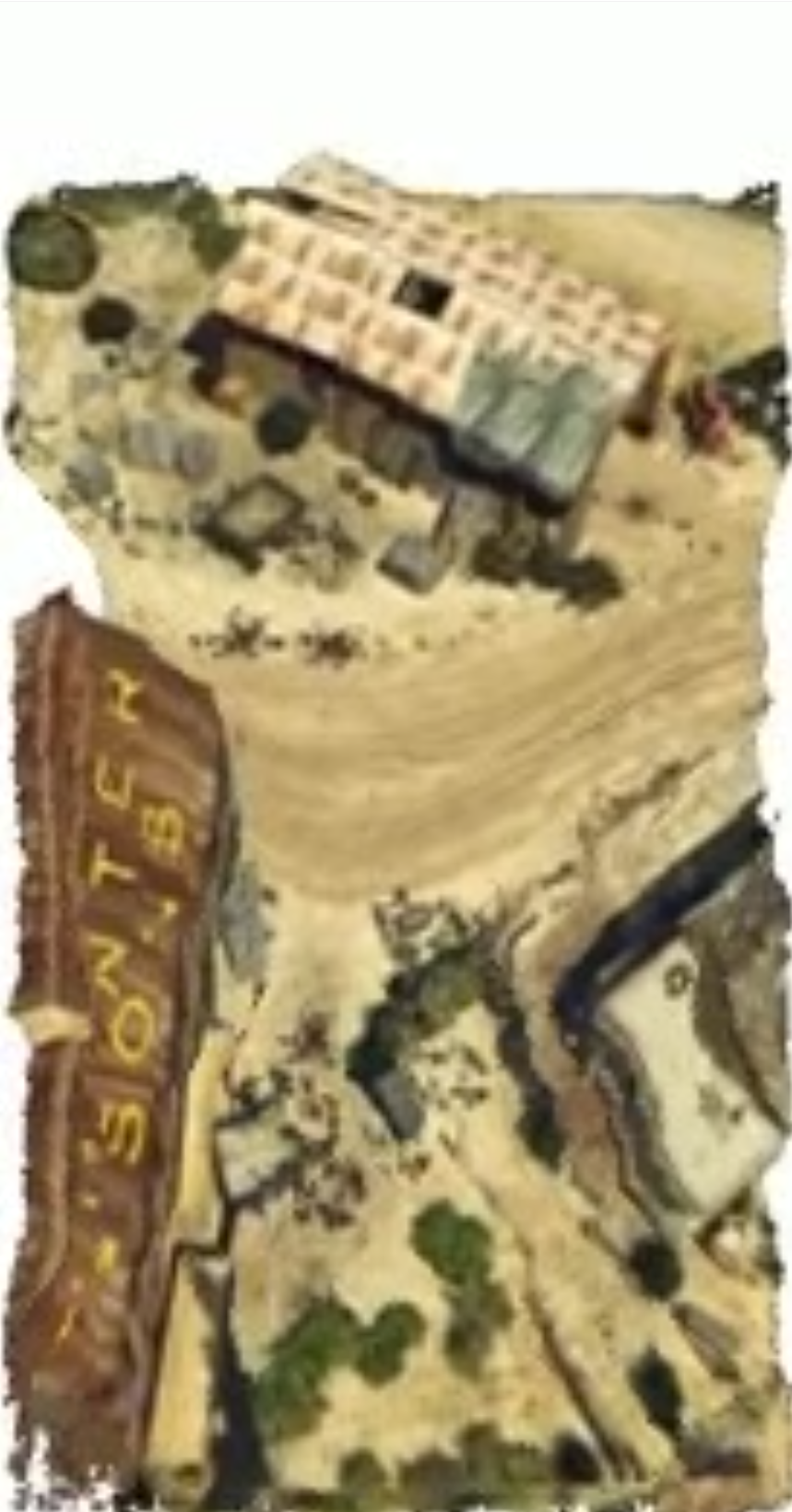}
    \end{tabular}
    \end{center}
    \captionof{figure}{Examples of our deformation approach compared to video seam carving \cite{video_seam}. Note the wrong perspective distortion that is introduced by video seam carving and the deformation of the umbrella or towels.}\label{table:qualitative_vsc}
\end{table*}
\newpage

\section{Additional Results}\label{app:extra_results}
All results in \cref{table:extra_image_results} use colour gradient, whereas \cref{table:extra_image_results_saliency} uses the saliency map from \cite{srinivas2019full} to improve results (see following pages). For a failure case, consider the last line: Our energy term allows expanding unconnected regions, hence expanding undesired regions that cause a bent in the trees. However, we observe that these failures occur more rarely than the failure modes of the other approaches.
\begin{table*}
    \begin{center}
    \begin{tabular}{ccccc}
Ours & Seam carving & Input & Seam carving & Ours\\
\includegraphics[height=0.1\linewidth]{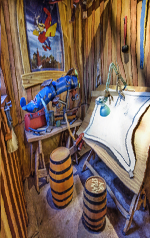}&\includegraphics[height=0.1\linewidth]{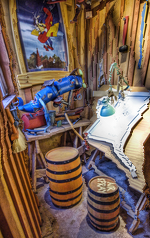}&\includegraphics[height=0.1\linewidth]{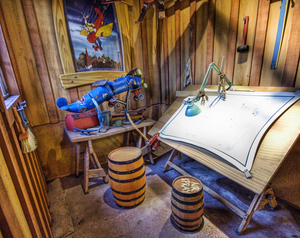}&\includegraphics[height=0.1\linewidth]{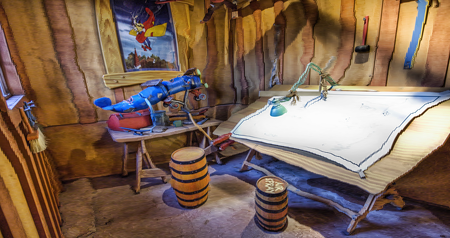}&\includegraphics[height=0.1\linewidth]{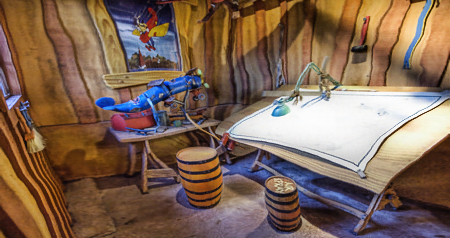}\\
\includegraphics[height=0.1\linewidth]{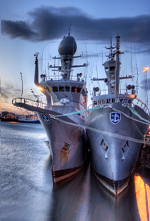}&\includegraphics[height=0.1\linewidth]{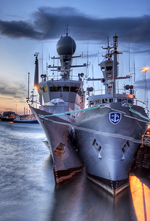}&\includegraphics[height=0.1\linewidth]{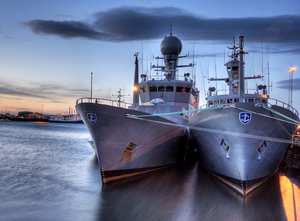}&\includegraphics[height=0.1\linewidth]{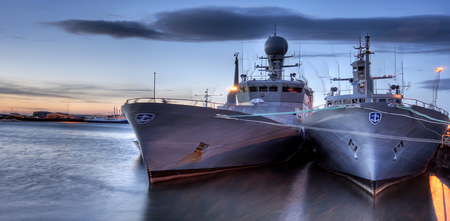}&\includegraphics[height=0.1\linewidth]{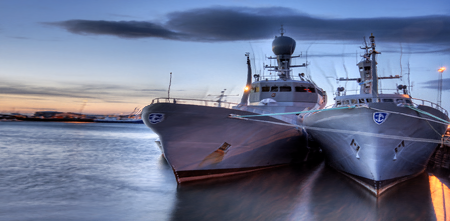}\\
\includegraphics[height=0.1\linewidth]{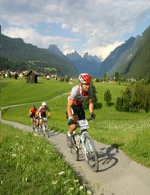}&\includegraphics[height=0.1\linewidth]{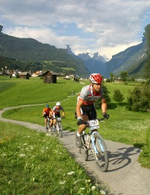}&\includegraphics[height=0.1\linewidth]{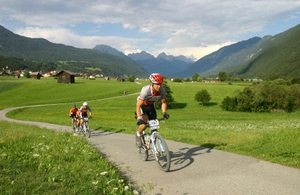}&\includegraphics[height=0.1\linewidth]{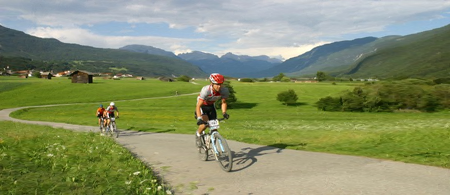}&\includegraphics[height=0.1\linewidth]{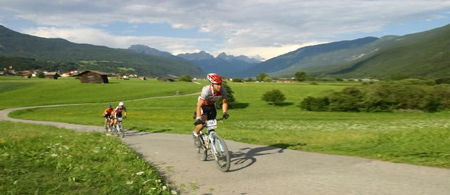}\\
Ours & Seam carving & Input & Seam carving & Ours\\
\includegraphics[height=0.1\linewidth]{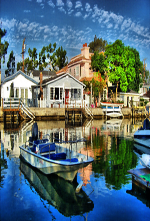}&\includegraphics[height=0.1\linewidth]{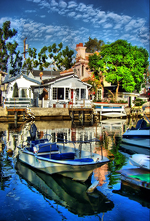}&\includegraphics[height=0.1\linewidth]{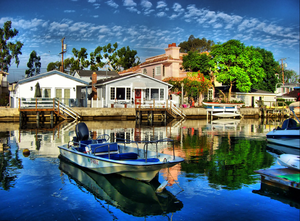}&\includegraphics[height=0.1\linewidth]{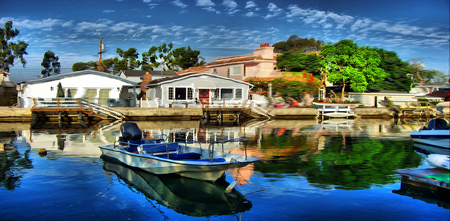}&\includegraphics[height=0.1\linewidth]{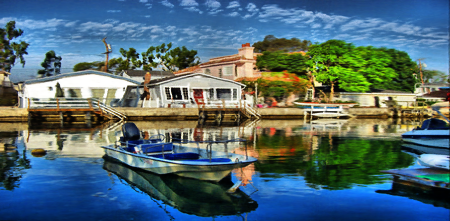}\\
\includegraphics[height=0.1\linewidth]{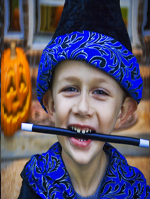}&\includegraphics[height=0.1\linewidth]{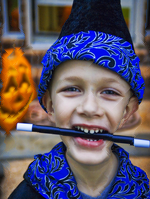}&\includegraphics[height=0.1\linewidth]{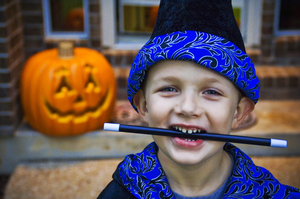}&\includegraphics[height=0.1\linewidth]{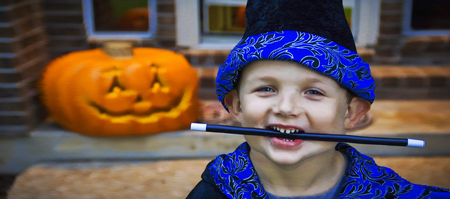}&\includegraphics[height=0.1\linewidth]{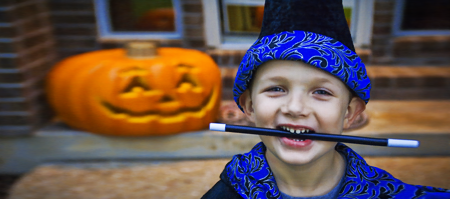}\\
\includegraphics[height=0.1\linewidth]{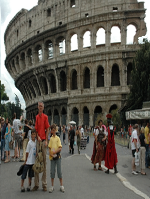}&\includegraphics[height=0.1\linewidth]{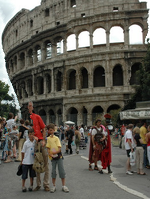}&\includegraphics[height=0.1\linewidth]{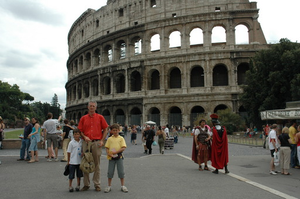}&\includegraphics[height=0.1\linewidth]{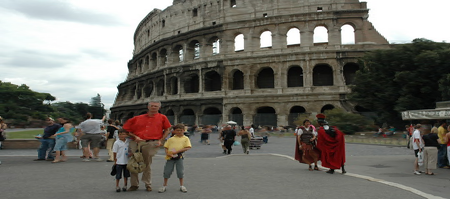}&\includegraphics[height=0.1\linewidth]{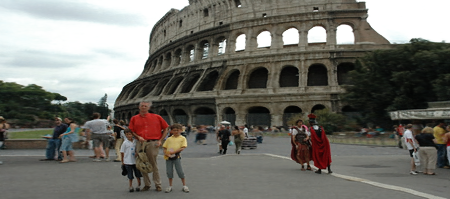}\\
Ours & Seam carving & Input & Seam carving & Ours\\
\includegraphics[height=0.1\linewidth]{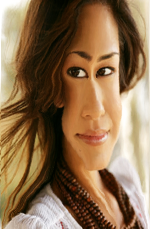}&\includegraphics[height=0.1\linewidth]{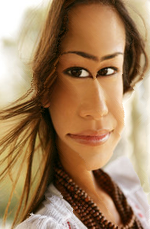}&\includegraphics[height=0.1\linewidth]{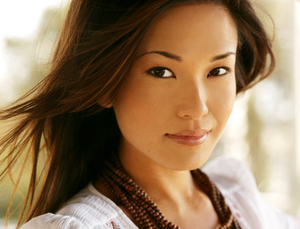}&\includegraphics[height=0.1\linewidth]{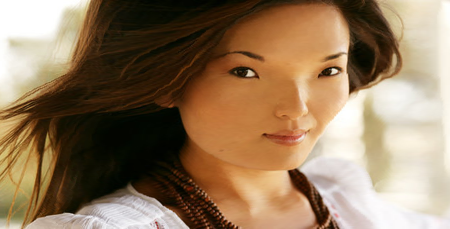}&\includegraphics[height=0.1\linewidth]{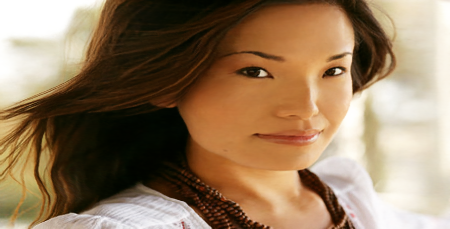}\\
\includegraphics[height=0.1\linewidth]{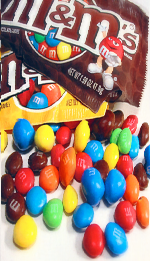}&\includegraphics[height=0.1\linewidth]{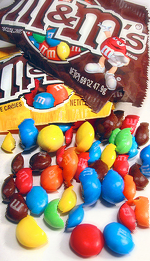}&\includegraphics[height=0.1\linewidth]{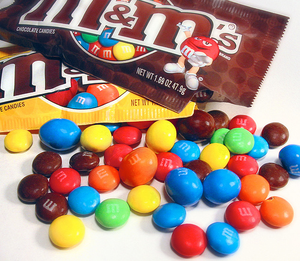}&\includegraphics[height=0.1\linewidth]{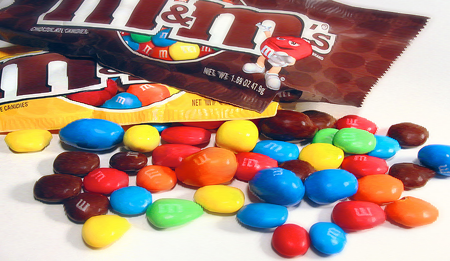}&\includegraphics[height=0.1\linewidth]{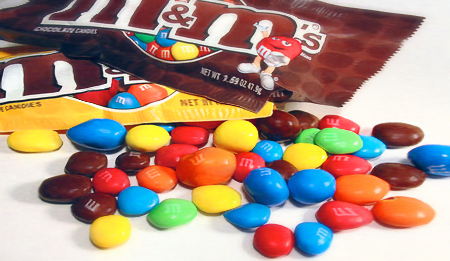}\\
\includegraphics[height=0.1\linewidth]{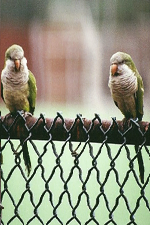}&\includegraphics[height=0.1\linewidth]{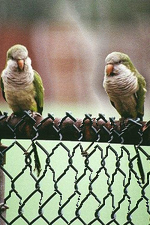}&\includegraphics[height=0.1\linewidth]{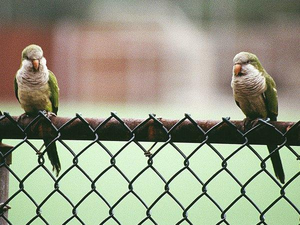}&\includegraphics[height=0.1\linewidth]{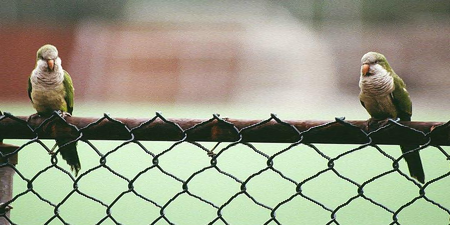}&\includegraphics[height=0.1\linewidth]{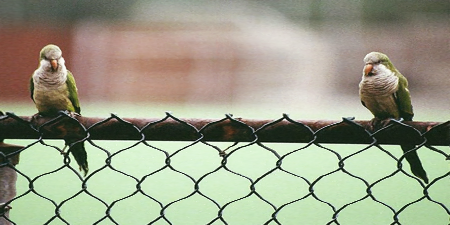}\\
Ours & Seam carving & Input & Seam carving & Ours, failure case below\\
\includegraphics[height=0.1\linewidth]{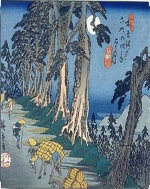}&\includegraphics[height=0.1\linewidth]{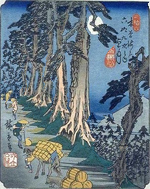}&\includegraphics[height=0.1\linewidth]{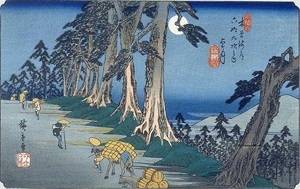}&\includegraphics[height=0.1\linewidth]{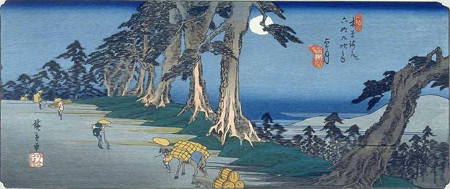}&\includegraphics[height=0.1\linewidth]{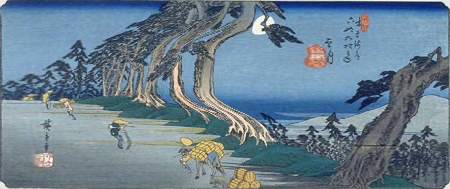}\\
    \end{tabular}
    \end{center}
    \captionof{figure}{Problematic examples of our deformation approach on images using colour gradient, all from the RetargetMe dataset~\cite{retargetme}.}\label{table:extra_image_results}
\end{table*}

\begin{table*}
    \begin{center}
    \begin{tabular}{cccc}
 &  & Ours, & Ours, \\
Input & Seam carving & colour gradient & saliency of \cite{srinivas2019full}\\
\includegraphics[width=0.3\linewidth]{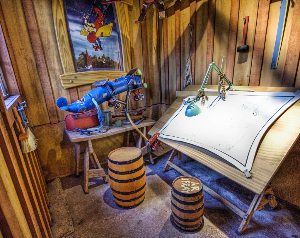}&
\includegraphics[width=0.15\linewidth]{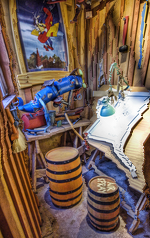}&
\includegraphics[width=0.15\linewidth]{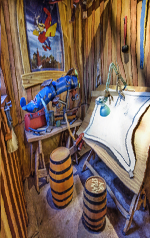}&
\includegraphics[width=0.15\linewidth]{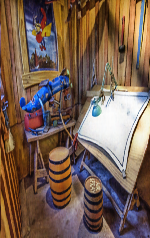}\\

\includegraphics[width=0.3\linewidth]{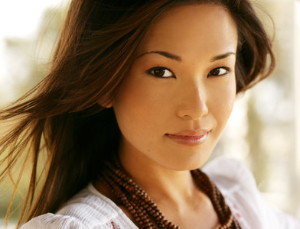}&
\includegraphics[width=0.15\linewidth]{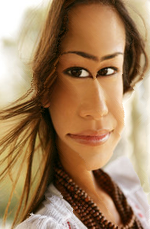}&
\includegraphics[width=0.15\linewidth]{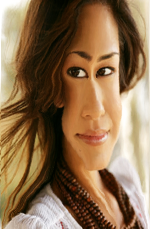}&
\includegraphics[width=0.15\linewidth]{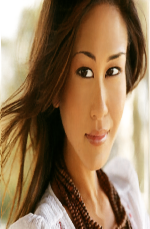}\\

\includegraphics[width=0.3\linewidth]{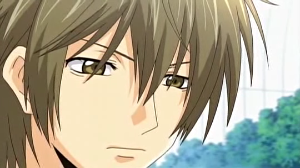}&
\includegraphics[width=0.15\linewidth]{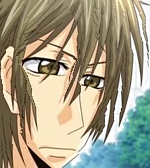}&
\includegraphics[width=0.15\linewidth]{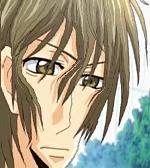}&
\includegraphics[width=0.15\linewidth]{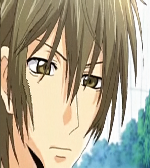}\\

\includegraphics[width=0.3\linewidth]{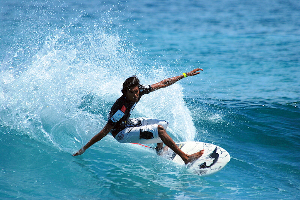}&
\includegraphics[width=0.15\linewidth]{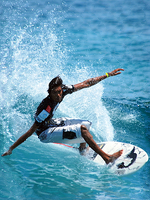}&
\includegraphics[width=0.15\linewidth]{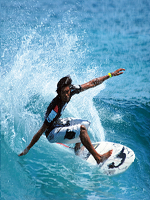}&
\includegraphics[width=0.15\linewidth]{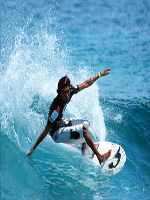}\\

\includegraphics[width=0.3\linewidth]{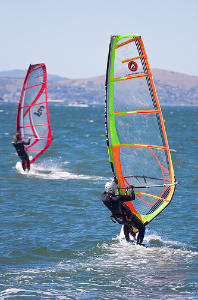}&
\includegraphics[width=0.15\linewidth]{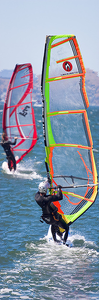}&
\includegraphics[width=0.15\linewidth]{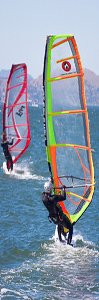}&
\includegraphics[width=0.15\linewidth]{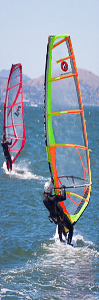}\\
    \end{tabular}
    \end{center}
    \captionof{figure}{Results compared to using the approach of \cite{srinivas2019full} as saliency map. All images from the RetargetMe dataset~\cite{retargetme}.}\label{table:extra_image_results_saliency}
\end{table*}

\begin{table*}
    \begin{center}
    \begin{tabular}{ccc}
    Ours to 50 \% & Input scene & Ours to 150 \%\\
\includegraphics[height=0.25\linewidth]{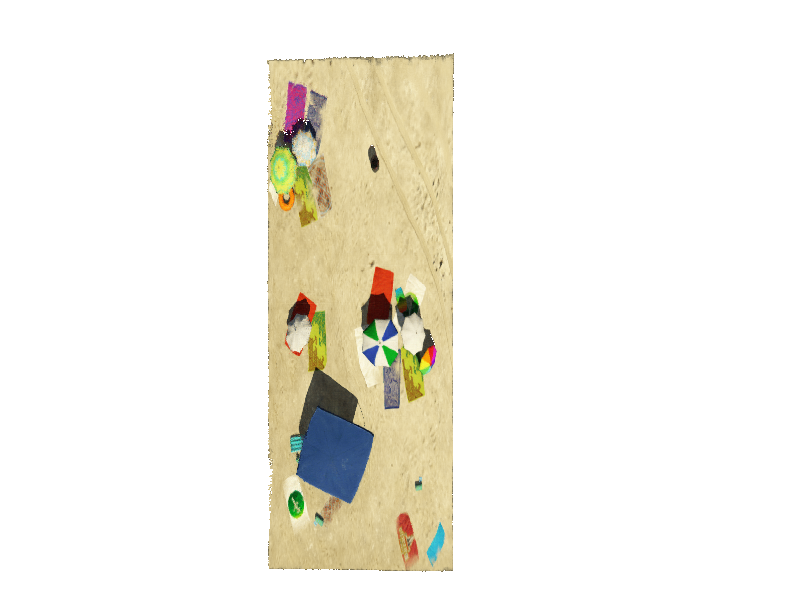}& \includegraphics[height=0.25\linewidth]{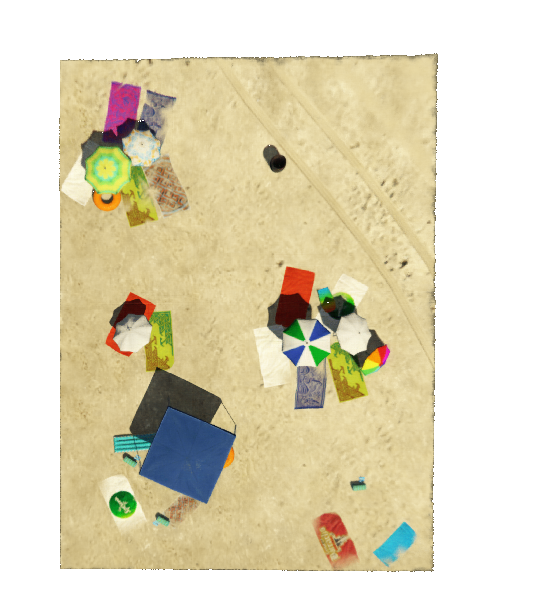}& \includegraphics[height=0.25\linewidth]{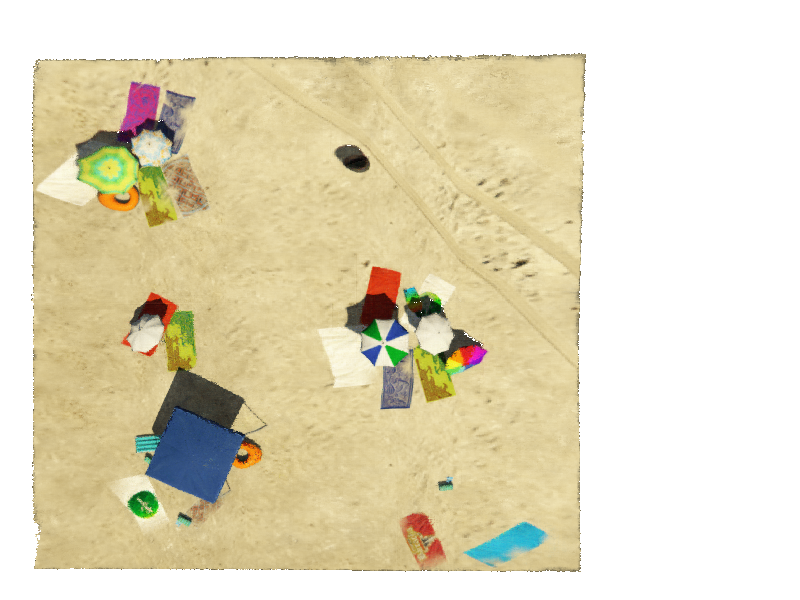}\\
\includegraphics[width=0.25\linewidth]{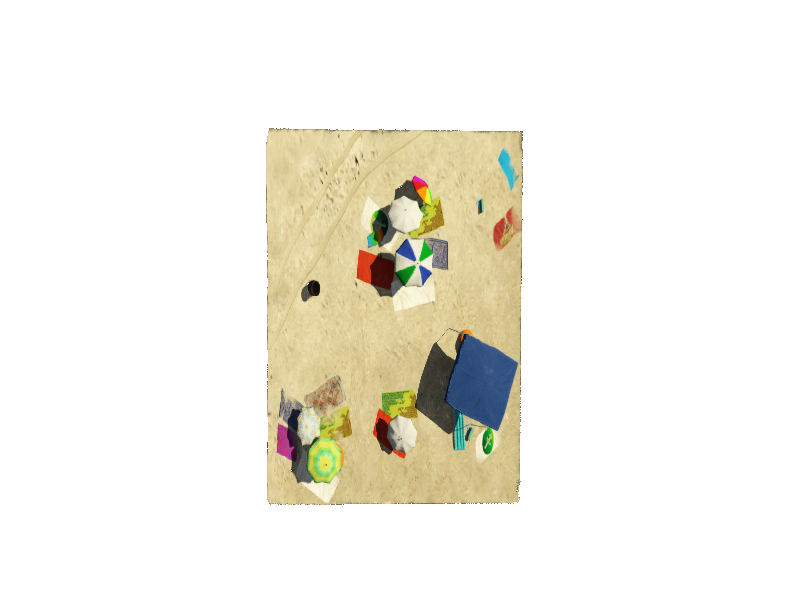}& \includegraphics[width=0.25\linewidth]{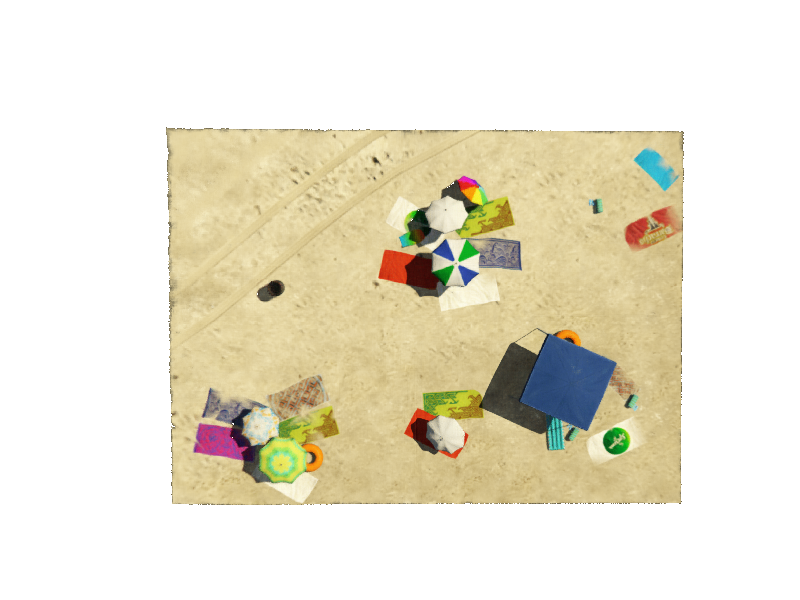}& \includegraphics[width=0.25\linewidth]{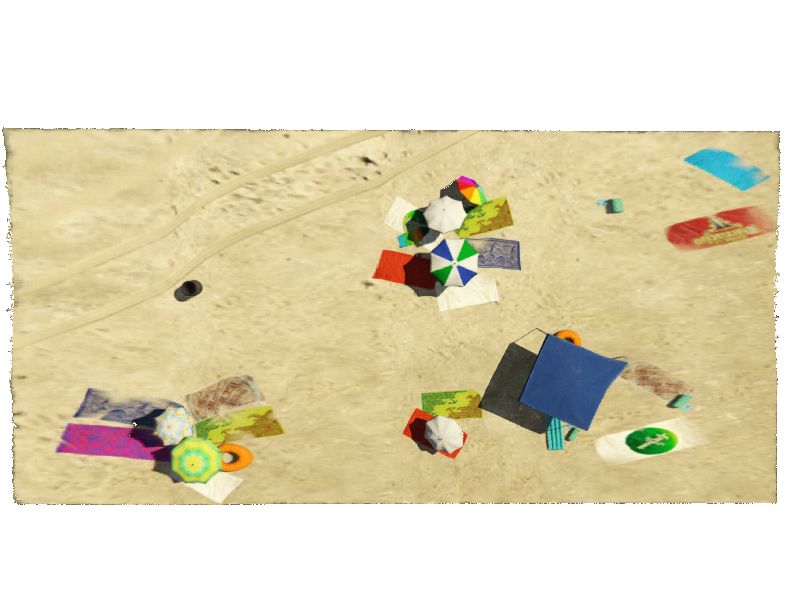}\\
\includegraphics[width=0.25\linewidth]{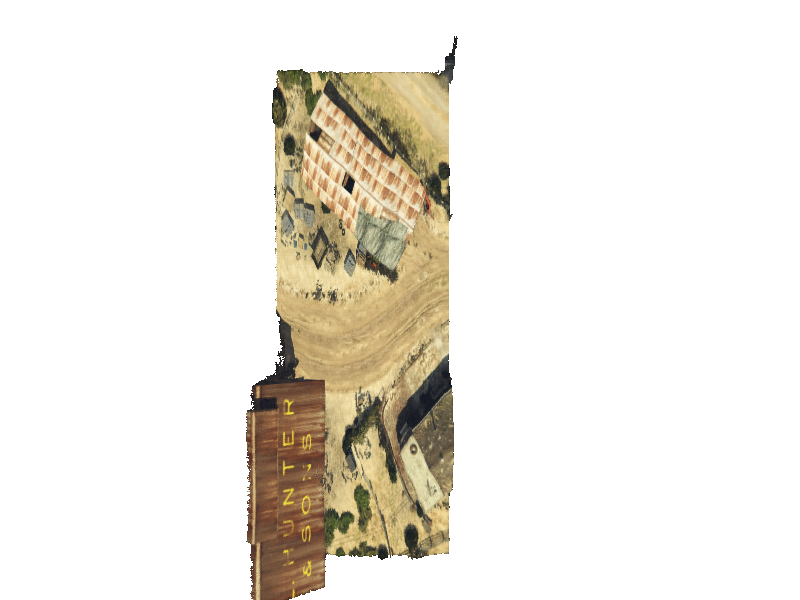}& \includegraphics[width=0.25\linewidth]{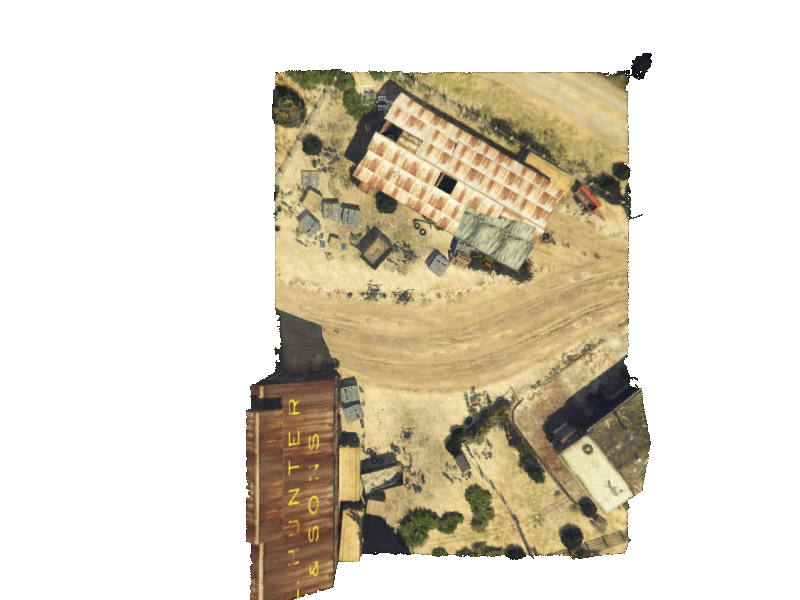}& \includegraphics[width=0.25\linewidth]{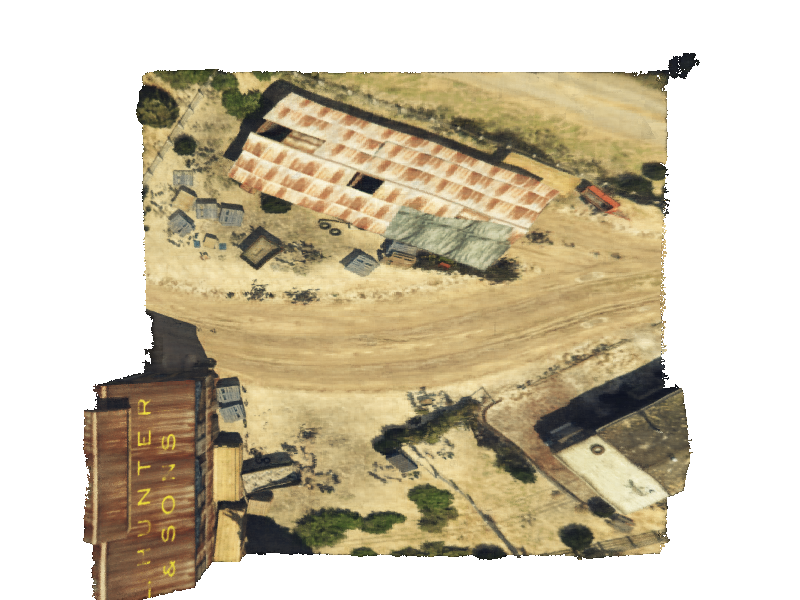}\\
    Ours to 50 \% & Input scene & Ours to 150 \%\\
\includegraphics[width=0.25\linewidth]{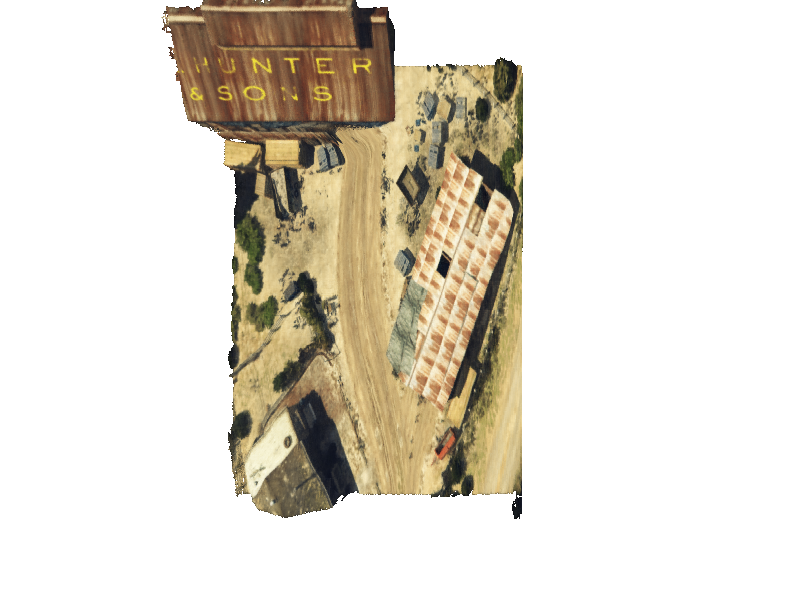}& \includegraphics[width=0.25\linewidth]{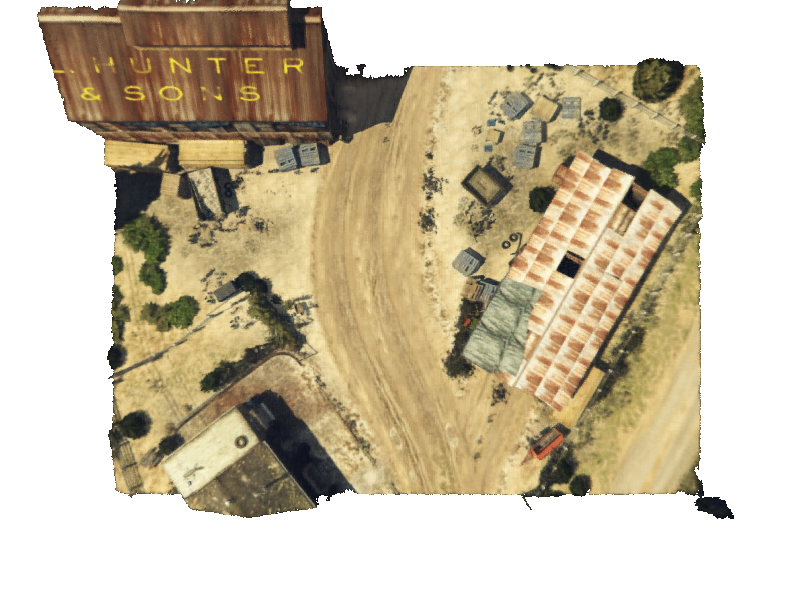}& \includegraphics[width=0.25\linewidth]{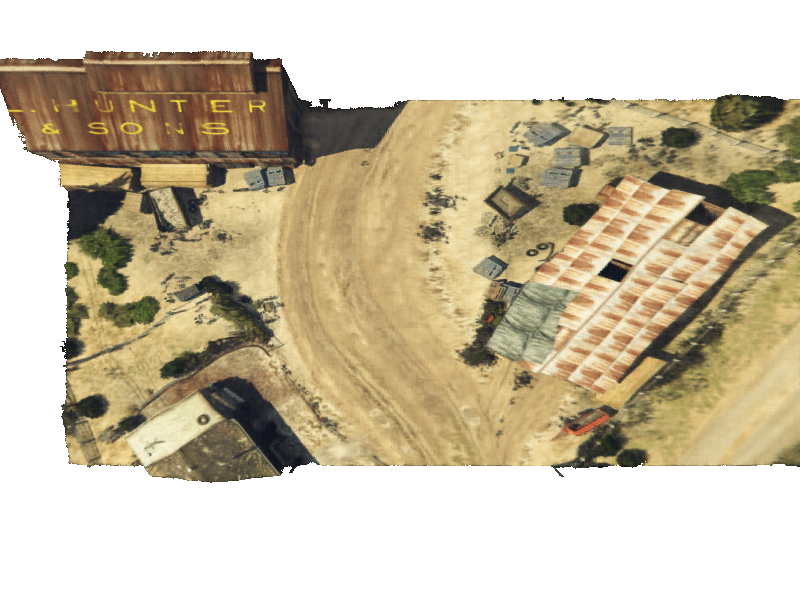}\\
\includegraphics[width=0.25\linewidth]{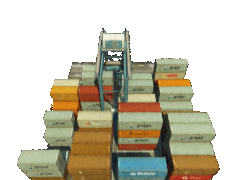}& \includegraphics[width=0.25\linewidth]{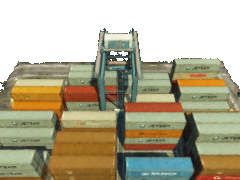}& \includegraphics[width=0.25\linewidth]{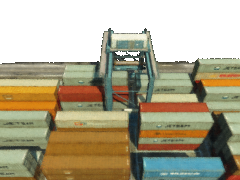}\\
\includegraphics[width=0.25\linewidth]{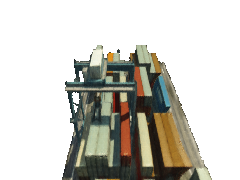}& \includegraphics[width=0.25\linewidth]{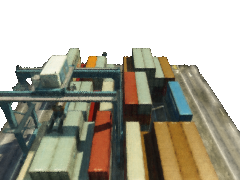}& \includegraphics[width=0.25\linewidth]{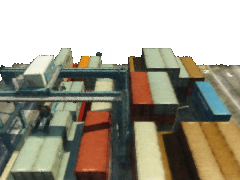}
    \end{tabular}
    \end{center}
    \captionof{figure}{More examples of our deformation approach on 3D NeRF scenes.}\label{table:extra_nerf_results}
\end{table*}

\end{document}